\documentclass[conference]{IEEEtran}
\usepackage{mathtools}
\usepackage{amsmath} 
\usepackage{amssymb}  
\usepackage{footnote}
\usepackage{graphics}
\usepackage{subcaption}
\usepackage{enumerate}
\usepackage{cite}

\ifCLASSINFOpdf
\else
\fi
\usepackage{algorithm}
\usepackage{algorithmic}
	\floatname{algorithm}{Algorithm}

\begin{document}
%
\title{A Coordinated Search Strategy for Multiple Solitary Robots: An Extension}

 \author{\IEEEauthorblockN{Jordan F. Masakuna}
\IEEEauthorblockA{\small Computer Science Division\\ CSIR-SU Centre for Artificial Intelligence Research \& \\
African Institute for Mathematical Sciences \\	
Stellenbosch University, South Africa\\
Email: jordan@aims.ac.za}
\and
\IEEEauthorblockN{Simukai W. Utete}
\IEEEauthorblockA{\small African Institute for Mathematical Sciences\\
Cape Town, South Africa\\
Email: simukai@aims.ac.za}
\and
\IEEEauthorblockN{Steve Kroon}
\IEEEauthorblockA{\small Computer Science Division\\
Stellenbosch University, South Africa\\
Email: kroon@sun.ac.za}}


%


\maketitle

\begin{abstract}
The problem of coordination without \textit{a priori} information about the environment is important in robotics. Applications vary from formation control to search and rescue. This paper considers the problem of search by a group of solitary robots: self-interested robots without \textit{a priori} knowledge about each other,  and with restricted communication capacity. When the capacity of robots to communicate is limited, they may obliviously search in
overlapping regions (i.e. be subject to interference). Interference hinders robot progress, and strategies have been proposed in the literature to mitigate interference \cite{IEEEexample:wellman2011using, IEEEexample:hourani2013serendipity}. Interaction of solitary robots has attracted much interest in robotics, but the problem of mitigating interference when time for search is limited remains an important area of research.  We propose a coordination strategy based on the method of cellular decomposition \cite{IEEEexample:choset2001coverage} where we employ the concept of soft obstacles: a robot considers cells assigned to other robots as obstacles. 
The performance of the proposed strategy is demonstrated by means of simulation experiments.  
Simulations indicate the utility of the strategy in situations where a known upper bound on the search time precludes search of the entire environment.
\end{abstract}


%
\IEEEpeerreviewmaketitle

\section{Introduction}
\label{sec:introduction}

In recent years, researchers and engineers have been increasingly interested in coordinating systems of intelligent robots, with applications ranging from formation control to search and rescue.
The problem of coordination is important in robotics, for example, where a group of robots need to achieve some task \cite{IEEEexample:arkin2002line, IEEEexample:choset2001coverage, IEEEexample:feinerman2012collaborative,IEEEexample:hao2008cooperative, IEEEexample:kong2006distributed, IEEEexample:rekleitis2004limited, IEEEexample:roy2001collaborative, IEEEexample:solanas2004coordinated, IEEEexample:wurm2008coordinated}.
This work considers the case of search by multiple robots.
Search of an environment consists of finding targets by exploring the environment. 
In this work, coverage is a proxy for finding targets: the exploration performance of a robot is taken to be proportional to its area covered. 

This paper considers situations where solitary robots: self-interested robots without \textit{a priori} knowledge about each other and limited communication, search an unknown environment. 
Real mobile robots have limited communication capacity \cite{IEEEexample:mosteo2008multi}.
A recent survey of solutions to problems of limited communication can be found in \cite{IEEEexample:amigoni2017multirobot}.

We assume that when a robot starts searching, the only information it has about the environment is the region that it can perceive from its initial location. The scale and boundary locations of the environment are also unknown, although a hard boundary can be sensed when within perception range. 
It is also assumed that robots have a limited time for search: the environment is wide and robots cannot explore it entirely, even were the environment known \cite{IEEEexample:wellman2011using}. The constraint on time could be, for example, due to limited battery power of robots. 

One of the main concerns induced when robots with limited communication  explore an unknown environment is interference: robots may obliviously search in overlapping regions which hinders robot progress. Interference can also be a concern for robots which some \textit{priori} knowledge  \cite{IEEEexample:wellman2011using, IEEEexample:hourani2013serendipity}.

In the case of solitary robots, the only moment that robots detect interference is when they meet. Under limited communication, the meeting of solitary robots is just a fortunate coincidence (as in accidental or symmetric rendezvous \cite{IEEEexample:alpern1995rendezvous}). In this work, the accidental rendezvous strategy will refer to a strategy where robots meet only accidentally.

When search involves multiple robots, one would like a group of robots that know about each other to explore non-overlapping regions -- this is termed sustainability in \cite{IEEEexample:hourani2013serendipity}. It is assumed that robots are effective in making use of information they possess. Consequently, each group member typically should cover almost the same area in size. (We ignore gradient in the exploration of regions.) 

This paper proposes a coordination strategy based on the method of cellular decomposition \cite{IEEEexample:choset2001coverage} and the use of soft obstacles. The proposed strategy provides sustainable performance when the time for search is limited (insufficient to cover the entire search area) and known. A shorter version of this work can be found in \cite{IEEEexample:masakuna2019coordinated}. Here we extend the theoretical and practical investigation of the proposed coordination strategy. 

%


The rest of the paper is organised as follows: Section \ref{sec:relatedwork} discusses related work, focusing particularly on the accidental rendezvous and periodic rendezvous strategies. 
Section \ref{sec:softobstacle} motivates and presents the proposed strategy.
Section \ref{sec:experiments} compares the approaches and discusses the results. 
In Section \ref{sec:conclusion}, we conclude and propose further work.

\section{Related Work}
\label{sec:relatedwork}
The accidental rendezvous strategy applies a frontier-based approach for exploration of unknown environments \cite{IEEEexample:yamauchi1997frontier, IEEEexample:yamauchi1998frontier}.  Frontiers
are  regions  at  the  boundary  between explored and unexplored regions. Frontier-based strategies are greedy algorithm and differ primarily in how frontiers are selected.

Greedy algorithms are
optimal for coverage problems, as they are a special form of
submodular optimization \cite{IEEEexample:nemhauser1978analysis}.

The main work using rendezvous in exploring unknown environments can be found in \cite{IEEEexample:wellman2011using, IEEEexample:hourani2013serendipity, IEEEexample:de2010selection}.
An interesting approach which provides a solution to mitigate interference with sustainable exploration performance was proposed in \cite{IEEEexample:wellman2011using}: at each rendezvous, the search space is divided into sectors (i.e. unbounded regions) to which robots are assigned. 
For example, the most
westerly part of the environment might be assigned to a robot. The soft obstacle strategy proposed here also applies this idea of assigning robots to sectors.

\begin{figure}[h]
	\centering
	\begin{subfigure}[h]{0.24\textwidth}
		\includegraphics[width=\textwidth]{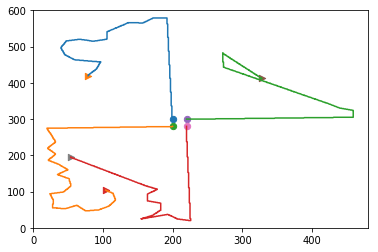}
		\caption{}
\label{fig:exampleARS1}
	\end{subfigure}
	\begin{subfigure}[h]{0.24\textwidth}
		\includegraphics[width=\textwidth]{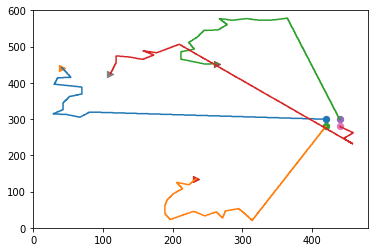}
			\caption{}
		\label{fig:exampleARS2}
	\end{subfigure}
	\caption{\textit{Illustration of assignment of sectors with four robots (coloured circles are robots, the lines show their respective trajectories).  \textbf{(\ref{fig:exampleARS1})}: even assignment, robots start at the centre of the environment. The chance of overlap is small. \textbf{(\ref{fig:exampleARS2})}: uneven assignment, robots ignore that they start close to a boundary, so the chance of overlap is high (see crossed lines).}}
	\label{fig:exampleARS}
\end{figure}

However, dividing areas into sectors can result in uneven assignment, which promotes interference. When uneven assignment, robots have different-sized areas to explore. 
For example, an uneven assignment arises when robots interact close to a boundary of the environment which they ignore while coordinating. In Figure \ref{fig:exampleARS2} for instance, the areas assigned to two of the robots (the green and red trajectories indicate these robots) are partly outside the environment, something the robots are unaware of when coordinating, as boundaries of the environment  are unknown until encountered.
Thus, assigning sectors under the accidental rendezvous strategy may lead to interference, and hence to non-sustainable performance. 

To address uneven assignment, robots could schedule further meetings to share additional information. However, scheduled meetings induce interruptibility: robots cease new knowledge acquisition when the rendezvous time arrives. They stop knowledge acquisition to travel to the rendezvous point. 
It has been shown that interruptibility can consume up to half of the search time \cite{IEEEexample:hourani2013serendipity}. 
 Methods exist which mitigate interruptibility.  For example, the exploration method proposed in \cite{IEEEexample:de2010selection} indirectly mitigated interruptibility.  
The method consists of two types of robots (explorer and relay) and a base station. The explorers intentionally meet and share knowledge with the relays. The latter transfer the shared information to the base station. Interruptibility is mitigated by the fact that the explorer and relay meet closer to the frontiers of the explorer. In the approach taken in this work, all robots are explorers and no base station is present.

Another solution in the literature which mitigates interruptibility has robots plan to meet other robots before the rendezvous time arrives \cite{IEEEexample:hourani2013serendipity}. Two challenging aspects need to be handled in this case: forecasting of the current positions of the robots involved; and forecasting of the paths taken by the robots from their current positions.  To address these challenges, the method proposed by Hourani \textit{et al.} is based on serendipity \cite{IEEEexample:hourani2013serendipity}. Usually, when robots interrupt exploration to attend a rendezvous, new information might not be gained. To address this negative impact, some robots, which are referred to as serendip robots, plan to interact with other members before the rendezvous time arrives by forecasting their paths. Thus interruptibility is mitigated.  

For the contribution here, this work applies the cellular decomposition strategy to search of an unknown environment by solitary robots. The approach aims to produce an interruptibility-free strategy with sustainable exploration performance.  Cellular decomposition allows robots to explore non-overlapping regions, thus interference is mitigated. Cellular decomposition also allows robots to avoid the need to schedule further rendezvous, thus reducing interruptibility. 

The region assigned to a robot is termed an exploration region. Here, a robot considers exploration regions assigned to others as soft obstacles. Thus, it eschews those regions, although a robot can move through soft obstacles if it would otherwise be trapped because surrounded by explored regions.
\section{Soft-obstacle Strategy}
\label{sec:softobstacle}
The proposed coordination strategy is an heuristic technique, which means it seeks for a satisfactory solution. Solitary robots, being self-interested, do not have a common goal. Solitary robots work under bounded rationality. Due to limited knowledge of a solitary robot in terms of the tractability of the problem and information about others, a satisfactory solution may be needed \cite{IEEEexample:simon1956rational}.
This work assumes the following:
\begin{itemize}
	\item the environment is a static, wide and initially unknown $2$D gridworld, although boundaries can be detected when encountered;
	\item targets are uniformly and independently distributed.
	\item robots are solitary, autonomous, homogeneous and uniquely identifiable.  A noise-free system is considered, but uncertainty can be incorporated;
	\item there could be many robots searching, but coordination involves only a small number of robots at a time (coordination of up to ten robots will be considered here).
\end{itemize}

The key concern with sectors (i.e. unbounded regions) which this work aims to handle is that of the behaviour of robots in cases of uneven or unbalanced assignment. We propose a way to allow a robot to explore outside of its sector, taking into account other robot sectors, in order to reduce interference. 
It is to be noted that the main difference between the soft obstacle strategy and the strategy of sectors applied in \cite{IEEEexample:wellman2011using} is that the soft obstacle strategy assigns bounded regions to robots. Sectors are not bounded. 

In summary, the problem considered in this paper is the provision of a coordination strategy for multiple solitary robots. In the case studied here, robots can coordinate only when they are in the same vicinity. The strategy targets the situation where search time is known and bounded, for example, by an attribute of the robot such as available battery time.  

\subsection{Intuitive considerations in the soft obstacle strategy}
\begin{figure}[h]
	\centering
	\includegraphics[scale=0.21]{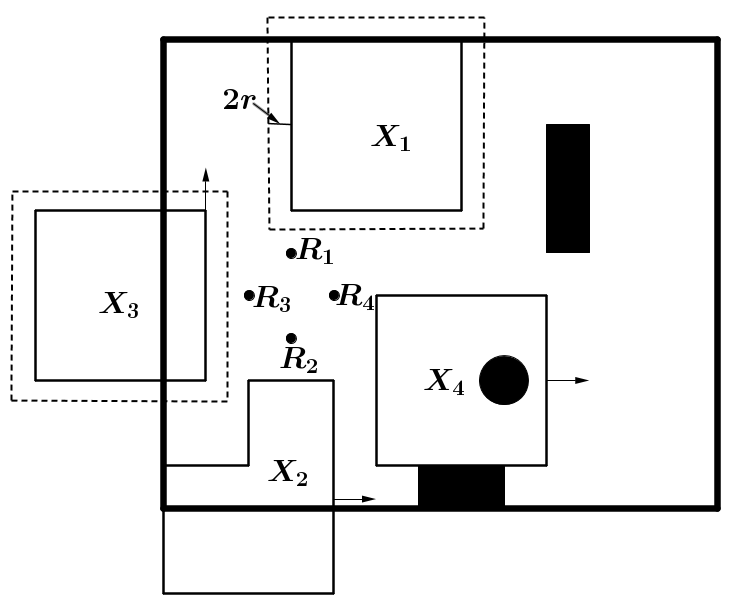}
	\caption{\small \textit{Illustration of a coordination based on the method of cellular decomposition with four robots. Black boxes and circle denote obstacles. The most thickened lines denote boundaries of the search space and the squares with solid lines are exploration regions.}}
	\label{fig:illustration}
\end{figure}

To contribute effectively to a group search, each robot should avoid searching areas others have already covered. To achieve this, we make use of the following two insights:
\begin{enumerate}
	\item since the upper bound on search time is known in advance, a robot can use this information to evaluate the maximum area it can cover;
	\item if a robot knows the regions assigned to other robots, it can plan to avoid exploring those regions. A robot is aware of other robots' regions only through rendezvous.
\end{enumerate}
These insights are used in our approach when constructing the exploration region points for the cellular decomposition strategy  \cite{IEEEexample:choset2001coverage}.
However, just as with the approach of sectors applied in \cite{IEEEexample:wellman2011using}, there remain other major concerns when applying cellular decomposition in an unknown environment. 
\begin{itemize}
	\item First, the region assigned to a robot could be partly outside the environment, which the robot is not aware of at the start ($X_3$ in Figure \ref{fig:illustration}). This is because robots might not have information about the boundaries of the environment while coordinating.
	\item Second, the exploration region of a robot could be occupied by obstacles (black boxes and circle in Figure \ref{fig:illustration}).
\end{itemize}

Areas assigned to a robot might include parts which are inaccessible because they are occupied by obstacles or because they lie outside the search region.  This can lead to robots having uneven assignments. When an uneven assignment is not addressed, robots may explore overlapping regions. Interference is more likely to occur in cases of uneven assignment. Let $\mathcal{W}$ and $\mathcal{O}$ denote the search environment and the set of obstacles in $\mathcal{W}$ respectively. With $\mathcal{W}^c$ the complement of $\mathcal{W}$, the inaccessible region is given by
\begin{equation}
\mathcal{O}\cup \mathcal{W}^c\,.
\end{equation}

A robot first creates a virtual world, meaning that some parts of its region might not be in the real environment. When its exploration region is partly accessible, the robot explores the accessible part first (details on exploring accessible regions will be provided later). Subsequently, the robot looks for another exploration region, taking into account the exploration regions assigned to others (i.e. soft obstacles) that it is aware of.  For a robot, the soft obstacles are prohibited regions. 

An important consequence of the above approach is that no matter what might prevent a robot from exploring its assigned exploration region (such as uneven assignment), other robots are prohibited from searching in that exploration region. This is an important feature of the strategy.

In Figure \ref{fig:illustration}, for example, most of the exploration region of robot $R_3$ is \textit{outside} the environment (i.e. it is inaccessible). The robot $R_3$, after exploring the accessible area of its exploration region, needs to get another exploration region. Given that $R_3$ knows the exploration regions of the other three robots, to get a fresh exploration region $R_3$ should follow a direction leading to the complement of the soft obstacles. In Figure \ref{fig:illustration} for example, $R_3$ can follow the direction indicated by the arrow on top of its exploration region. 

%

In coordination, the robots concerned choose a leader. To choose a leader, we consider the approach used in \cite{IEEEexample:kim2013leader}. In \cite{IEEEexample:kim2013leader}, a leader is elected based on closeness centrality\textemdash the closeness centrality of a robot is the reciprocal average of the depths between the robot and all other robots. Robots involved in interaction build a spanning tree first by message passing, then they choose a leader which has the highest closeness centrality. We consider situations where choosing a leader at the central position is important, for efficient transmission.

The elected leader assigns exploration regions to the participating robots using the Hungarian algorithm \cite{IEEEexample:kuhn1955hungarian} to minimize robot-to-region travelling time. 
Robots use straight lines to reach their respective exploration regions while avoiding obstacles or soft obstacles.

In addition to the problem of assignment of regions to robots, an obstacle avoidance system must be incorporated in order to enable robots to avoid obstacles encountered. Techniques used to avoid obstacles are introduced next.
\subsection{Obstacle avoidance techniques}
\label{sec:avoidance}
We consider two situations for obstacle avoidance: when a robot is moving to an exploration region and when a robot is exploring its region. For the former case, various methods can be applied which include the potential field, the vector field histogram methods and the local navigation method \cite{IEEEexample:khatib1986real, IEEEexample:ulrich1998vfh,IEEEexample:fujimori1997adaptive}. For simplicity, we use a Bug algorithm \cite{IEEEexample:oroko2014obstacle}. 
Bug algorithms are simple methods used for path planning to avoid an unexpected obstacle in the robot motion by updating the directional angle of a robot when an obstacle is detected.

Bug algorithms assume only local knowledge of the search environment (i.e. obstacles are unknown) and a goal, and the robot knows directions (and can evaluate the Euclidean distance) towards the goal. The robot has a limited perception range. When a robot detects an obstacle, it follows the trajectory of the obstacle boundary.
Bug algorithms are based on the two following principles. To avoid a detected obstacle the robot,
\begin{itemize}
	\item follow the closest edge of the obstacle, towards right, left, up or down.
	\item move in a straight line toward a goal.
\end{itemize} 
Here, goal can be either the exploration region of the robot (when the robot is travelling to its exploration region) or a boundary of its exploration region (when the robot is exploring its region using zigzag). 

There exist various variants of Bug algorithms, including Bug 1, Bug 2, and Distance Bug. Some Bug algorithms, as listed above, can be seen as improvement to others to some extent. For instance Distance Bug is an improvement of Bug 2 and the latter is an improvement to Bug 1. The three variants are described as follows.

\begin{figure}[h]
	\centering
	\begin{subfigure}[h]{0.24\textwidth}
		\includegraphics[width=\textwidth]{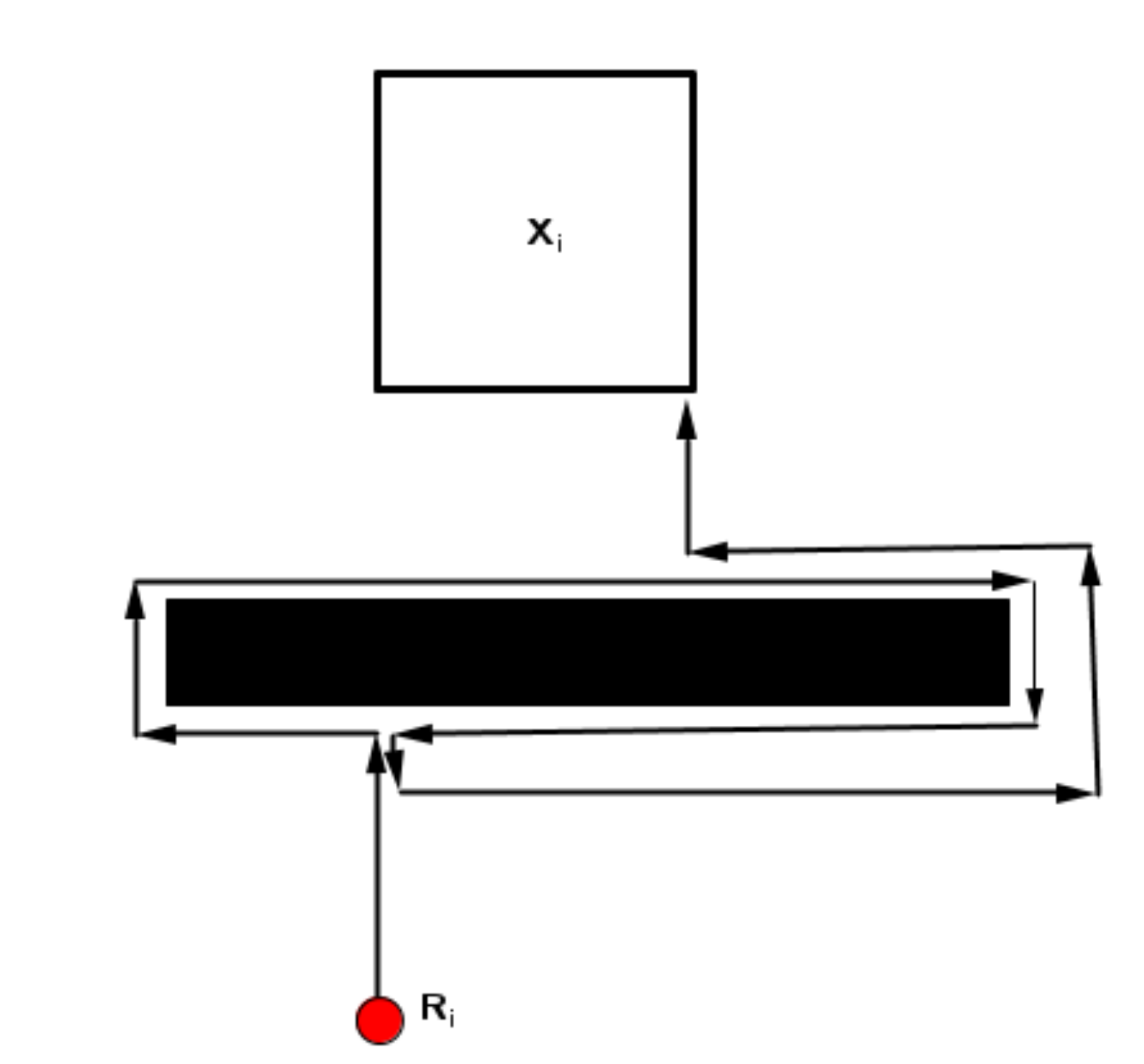}
		\caption{}
		\label{fig:bug1}
	\end{subfigure}
	\begin{subfigure}[h]{0.23\textwidth}
		\includegraphics[width=\textwidth]{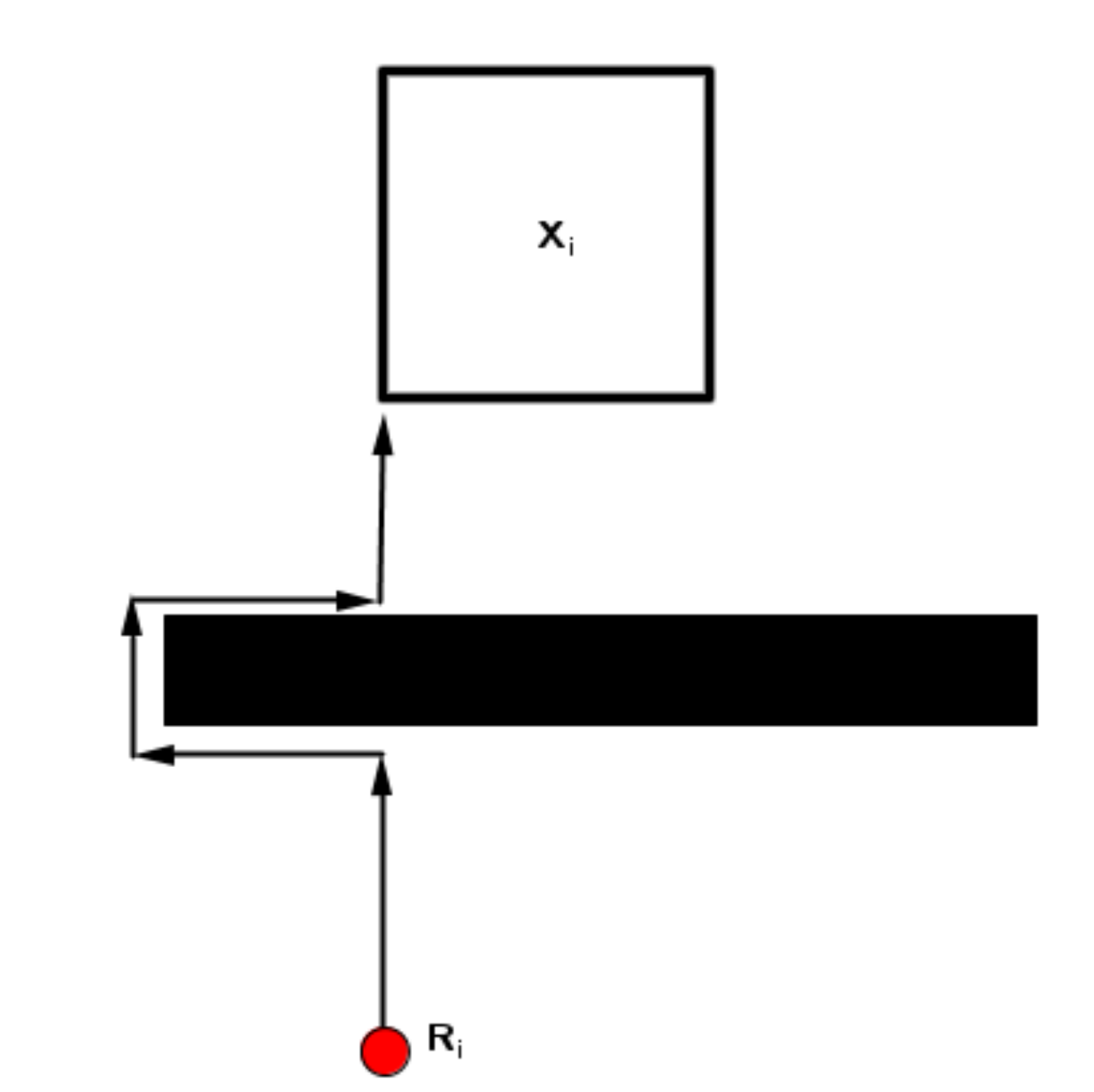}
		\caption{}
		\label{fig:bug2}
	\end{subfigure}
	\begin{subfigure}[h]{0.2\textwidth}
		\includegraphics[width=\textwidth]{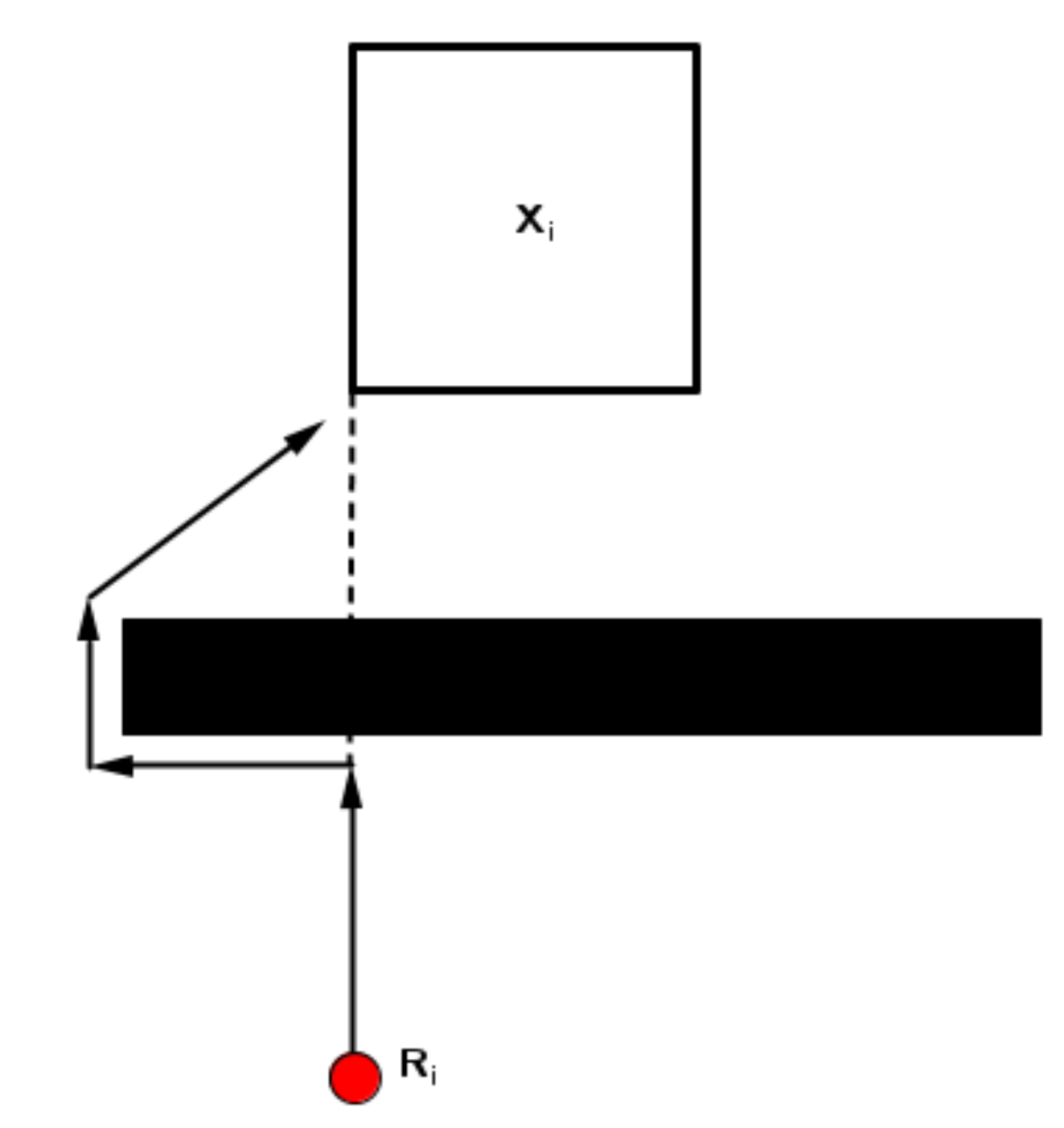}
		\caption{}
		\label{fig:bugdistance}
	\end{subfigure}
	\caption[\textit{Illustration of obstacle avoidance using Bug algorithms.}]{\textit{Illustration of obstacle avoidance using Bug algorithms where red filled circle denotes a robot $R_i$, black rectangle an obstacle and white square with black boundary the exploration region of $R_i$, $X_i$. \textbf{(\ref{fig:bug1})}: Bug 1. \textbf{(\ref{fig:bug2})}: Bug 2. \textbf{(\ref{fig:bugdistance})}: Distance Bug.}}
	\label{fig:bugs}
\end{figure}

In Bug 1, when a robot encounters an obstacle while travelling towards the goal, it follows a canonical direction until the location of the initial encounter is reached. After reaching the initial encounter, the robot follows the boundary of that obstacle to reach the point along the boundary that is closest to the goal. Once reached that closest point, the robot moves directly toward the goal. It repeats the same behaviour when new obstacles are encountered (Figure \ref{fig:bug1}). The revolving behaviour of the robot around each and every obstacle is computationally costly. Let $x_i$ and $a_i$ be the initial point of the robot and the goal respectively. Let $C(x_i, a_i)$ denote the performance of a robot to get to $a_i$ from $x_i$. The worst case performance is
upper bounded by
\begin{equation} C(x_i, a_i)\leq \delta(x_i, a_i) + \frac{3}{2} \sum_{j=1}^{k} p_j\,,
\end{equation}
where $\delta(x_i, a_i)$ is the Euclidean distance $x_i$ to $a_i$, $p_j$ is the perimeter of the $ij$th obstacle, and $k$ is the number of obstacles encountered by the robot. Bug 2 addresses this issue in Bug 1.

In Bug 2, when a robot encounters an obstacle, it starts moving along the edge of the obstacle until it finds a point along the boundary of the obstacle with the same slope (Figure \ref{fig:bug2}). Then the robot moves to the goal. The worst case performance is upper bounded by
\begin{equation} C(x_i, a_i)\leq\delta(x_i, a_i) + \frac{1}{2} \sum_{j=1}^{k} n_jp_j\,,
\end{equation}
where $n_j$ is the number of times the $j$th obstacle crosses the line segment between $x_i$ and $a_i$. In Bug 2, looking for a point along the boundary generating a line to the goal with the same slope as that of the line from $x_i$ and $a_i$ can slow the procedure of obstacle avoidance. Distance Bug, which we use in this work, improves Bug 2 in that regards. In Distance Bug, When a robot encounters an obstacle
, it chooses points which minimise the distance to the goal while avoiding obstacle (Figure \ref{fig:bugdistance}).  The worst case performance is upper bounded by
\begin{equation} C(x_i, a_i)\leq\delta(x_i, a_i) + \frac{1}{3} \sum_{j=1}^{k} n_jp_j\,,
\end{equation}
where $n_j$ is the number of times the $j$th obstacle crosses the line segment between $x_i$ and $a_i$.

Regarding obstacle avoidance during a robot's exploration of its assigned region, we propose a simple strategy to avoid obstacles. A robot applies a motion pattern (the motion pattern will be described in Subsection \ref{sec:zigzag}) to search its assigned region. If it encounters an obstacle, it attempts to maintain the pattern to the extent possible while avoiding the obstacle. This means that when it completes the current motion pattern some sub-regions might still be relatively unexplored. When it finishes applying its current motion pattern, it therefore must determine whether its current exploration region still contains significant unexplored areas, and to focus on these, or to seek a fresh search region based on the remaining search time.
%
To determine the unexplored area, a robot generates seeds over its exploration region, sampled from a uniform distribution. The seeds indicate the size and location of the unexplored area.
Figure \ref{fig:avoidance} and Figure \ref{fig:avoidance2} illustrate obstacle avoidance.
\subsection{Soft-obstacle strategy: cellular decomposition}
Let $X_i$, $\mathcal{C}^{(t)}_i$ and $\mathcal{Z}^{(t)}_i$ denote the exploration region, the explored region and the interference region of robot $R_i$ at time $t$ respectively.

Two components make up soft obstacles: the region $\mathcal{C}^{(t)}_i$ already explored and known to the robot $R_i$ and the interference regions $\mathcal{Z}^{(t)}_i$, i.e., exploration regions of other robots. The soft obstacles $\mathcal{B}^{(t)}_i$ of robot $R_i$ are
\begin{equation}
\label{eq:constructsoftobstacle}
\mathcal{B}^{(t)}_i = \mathcal{C}^{(t)}_i \cup \mathcal{Z}^{(t)}_i\,.
\end{equation} 
$R_i$'s exploration region $X_i$ is a grid and is composed of at least one cell. Initially, $X_i$ has a single cell. It is decomposed into other cells when $R_i$ avoids encountered obstacles in $X_i$ as explained above in Subsection \ref{sec:avoidance}. Assume that the coordination procedure lasts for $\Delta T$. At the end of coordination, the interference region of a robot $R_i$ is updated to include the other robots' exploration regions.
\begin{equation}
\label{eq:interferenceregion}
\mathcal{Z}^{(t+\Delta T)}_i=\mathcal{Z}^{(t)}_i\cup \bigg(\bigcup_{j\neq i}X_{j}\bigg)\,,
\end{equation}
An exploration region of a robot is characterised by a width, a height, and its left lower corner.
%

The sensing area by a robot is a circle. \footnote{But the sensing area can take other shapes. This should only have a slight impact on the initial and final conditions of the solution.}The maximum area that a robot can scan in $\tau$ time units is 
\begin{equation}
\label{eq:area}
A(\tau)= \pi \, r^2+2\gamma r\tau\,,
\end{equation}
where $\gamma$ ($\gamma>0$) denotes the motion scale and $r$ denotes the perception range of robots. A motion scale is a positive real number that denotes the number of points scanned on a line by a robot per unit
of time and it is fixed. Robots can have different motion scales. In this work homogeneous solitary robots have the same motion scale, $\gamma$. 
In interaction, robots fuse their individual information and assign to each other fresh exploration regions. Algorithms used to fuse data and to mission robots are described next.
\subsection{Data fusion and mission plans}
A local leader has three roles: it receives individual data from others, it coordinates the plans for their next actions and replicates decisions (missions) to others. Since search is the use case of this paper, robots' data also contains maps. Robots need to combine their individual maps.

The following step is to send individual data to the leader. 
Each robot has its own internal coordinate system. When two robots intend to merge their maps, unification of internal coordinate systems must be done. To fuse data, we consider a general framework proposed in \cite{IEEEexample:konolige2003map}.
The framework uses robot-to-robot measurements to achieve map fusion. 

The process of data fusion starts from robots with lower degrees. The leader must be the last to receive. 

A robot $R_i$  sends data to robot $R_j$ following format $(\mathcal{H}^{(t)}_i, \mathcal{C}^{(t)}_i)$ where $\mathcal{C}^{(t)}_i$ denotes the map from $R_i$ and $\mathcal{H}^{(t)}_i$ the interaction history of $R_i$ at time $t$ from their respective coordinate systems. 
The algorithm of data fusion applied by a single robot $R_i$ is illustrated in Algorithm \ref{algo:sending}. In Algorithm \ref{algo:sending}, $\mathcal{N}_i$ denote the immediate neighbours of the robot $R_i$.
\begin{algorithm}
	\begin{algorithmic}[1]
		\scriptsize
		\FORALL{$R_j\in \mathcal{N}_i$ such that $deg(R_j)< deg(R_i)$}
		\STATE $\mathcal{C}^{(t)}_i\gets  \mathcal{C}^{(t)}_i \cup \mathcal{C}^{(t)}_j $
		\STATE $\mathcal{H}^{(t)}_i \gets \mathcal{H}^{(t)}_i  \cup \mathcal{H}^{(t)}_j  $
		\ENDFOR
	\end{algorithmic}  
	\caption{data-fusion()}
	\label{algo:sending}
\end{algorithm}

When the leader has received all the information, it applies the coordination strategy. 
After combining individual data and making plans, the leader replicates the results to others. The algorithm that a robot $R_i$ applies for mission plans is given in Algorithm \ref{algo:reply}. The involved robots adopt the coordinate system of the leader. In Algorithm \ref{algo:reply}, $(XY)_i$ denotes the internal coordinate system of the robot $R_i$.
\begin{algorithm}
	\begin{algorithmic}[1]
		\scriptsize
		\FORALL{$R_j \in \mathcal{N}_i$ and $deg(R_j)>deg(R_i)$ }
		\STATE	$\mathcal{H}^{(t)}_j  \gets \mathcal{H}^{(t)}_i $
		\STATE	$(XY)_j  \gets (XY)_i $
		\STATE	gets the exploration region $X_j$ of $R_j$
		\ENDFOR
	\end{algorithmic} 
	\caption{mission-plan()}
	\label{algo:reply}
\end{algorithm}

The interaction history of $R_i$ is given as follows,
\begin{equation}
\label{eq:history}
\mathcal{H}^{(t)}_i = \{(R_j, l_j, h_{j}, a_j):\, \forall R_j\in \mathcal{N}_i\}\,,
\end{equation}
where $R_j$ denotes robots already interacted with $R_i$ (or robots which $R_i$ has histories of), $l_{j}$, $h_{j}$ and $a_j$ denote length, height and centroid of the exploration region of the robot $R_j$. 

In applying cellular decomposition in the proposed coordination strategy, the elected leader intentionally leaves unexplored spaces (or margins) between robots' exploration regions (the dotted blue lines around an exploration region in Figure \ref{fig:illustration}). Margins are to be used by robots whose exploration regions are partly inaccessible as each robot plans independently. Use of margins is part of the novelty of the proposed approach.
\subsection{Measure of margins}
\label{subsec:decomposition}
 When the region $X_i$ of $R_i$ is partly accessible, the margin $M_i$ will allow the robot $R_i$ to avoid interference when it is travelling to a new exploration region. It should be noted that robots also explore margins.  Once in a margin, the robot considers the margin as an exploration region and explores it entirely before moving to its fresh exploration  region. In other words, a margin is a path from the current location of the robot to its new exploration region. The area of the margin $M_i$ of a robot $R_i$ is proportional to a time $\tau_0$. The value of $\tau_0$, given in Equation \ref{eq:tau0}, is the amount of time a robot will take to explore its margin. 
The width of a margin is $m$ ($m\geq 2r$). 
\begin{equation}
\label{eq:tau0}
\tau_0=\frac{2}{r\gamma} \bigg(m^2 + m\sqrt{\pi r^2+2\gamma r (\tau-t-\Delta T)}\bigg)\,.
\end{equation}

The augmented exploration region of $R_i$, which is $M_i\cup X_i$, has the following area
\begin{equation}
\label{eq:newarea}
\pi \, r^2+2\gamma r(\tau+\tau_0-t-\Delta T)\,,
\end{equation}
where $t$ is the amount of time already spent.
The approach used by a robot to explore its region is discussed next.
\subsection{Zigzag search}
\label{sec:zigzag}
\begin{figure}[h]
	\centering
	\begin{subfigure}[h]{0.12\textwidth}
		\includegraphics[width=\textwidth]{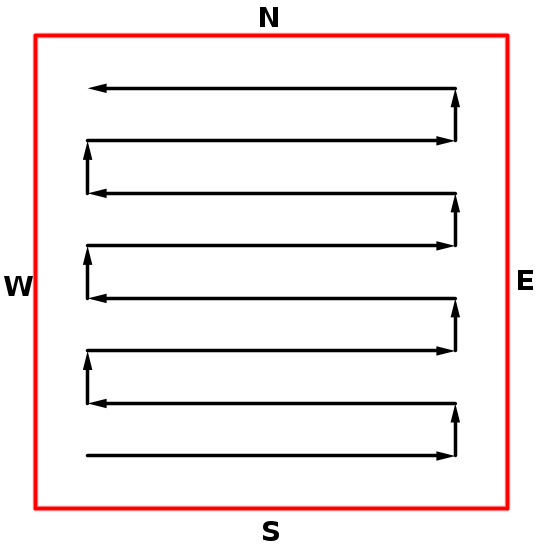}
		\caption{}
		\label{fig:descrzigzag1}
	\end{subfigure}
	\begin{subfigure}[h]{0.12\textwidth}
		\includegraphics[width=\textwidth]{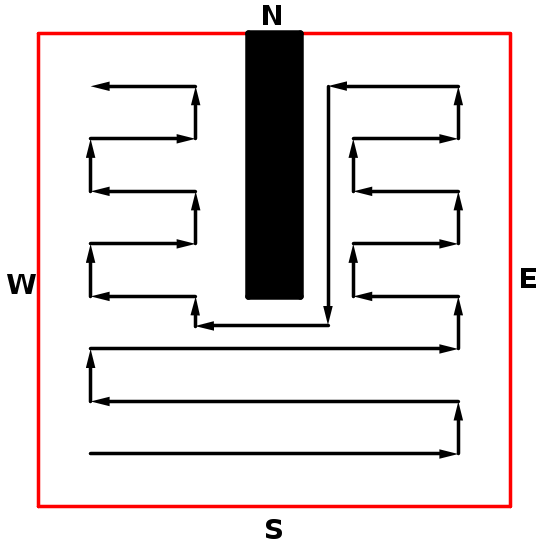}
		\caption{}
		\label{fig:avoidance}
	\end{subfigure}
	\begin{subfigure}[h]{0.12\textwidth}
		\includegraphics[width=\textwidth]{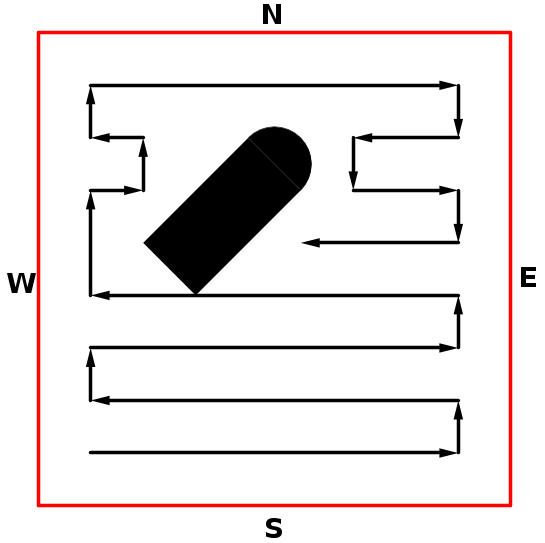}
		\caption{}
		\label{fig:avoidance2}
	\end{subfigure}
	\caption{\small \textit{Application of zigzag search. \textbf{(\ref{fig:descrzigzag1})}: an obstacle-free exploration region. \textbf{(\ref{fig:avoidance2})}: a region occupied by a regular polygonal obstacle. \textbf{(\ref{fig:avoidance})}: a region occupied by a non-regular polygonal obstacle.}}
	\label{fig:zigzag}
\end{figure}

To cover its exploration region, a robot applies a zigzag search\footnote{But other algorithms could be applied such as the spiral search and frontier-based search.} \cite{IEEEexample:choset2001coverage}. For the benchmark coordination algorithms, a robot will use a frontier-based search algorithm to cover its exploration region.
In zigzag search, the robot moves back and forth from one side of its region to the other, along parallel paths. For instance, a robot may move from west to east back and forth towards the north (Figure \ref{fig:descrzigzag1}).
%
%
%
The soft obstacle strategy used by a robot $R_i$ is given in Algorithm \ref{algo:softobstacle}.

\begin{algorithm}
	\begin{algorithmic}[1]
		\scriptsize
		\STATE $t\gets 0$
		\STATE the location $x_i$ of $R_i$ is set to its initial position
		\STATE $\mathcal{C}^{(t)}_i \gets B_r(x_i), \mathcal{Z}^{(t)}_i \gets \emptyset$
		\STATE initialise the exploration region $X_i$ from Equation \ref{eq:area}
		\STATE $R_i$ applies the zigzag search
		\WHILE{$t<\tau$ and unexplored regions available}
		\IF{$R_i$ interacts with $P$ $(P\geq 1)$ robot(s)}
		\STATE choose a leader $R_k$ among the interacting robots
		\STATE $R_k$ combines the explored region $\mathcal{C}^{(t)}_j$ and interaction histories $\mathcal{H}^{(t)}_j$ of other robots $R_j$ using Algorithm \ref{algo:sending}
		\IF{the robot $R_i$ is the elected leader}
		\STATE $R_i$ assigns mission plans to other robots using Algorithm \ref{algo:reply}
		\ELSE
		\STATE $R_i$ waits for its exploration region from the leader
		\ENDIF
		\STATE $R_i$ updates its soft obstacle $\mathcal{B}^{(t+\Delta T)}_i$
		\STATE $t\gets t+\Delta T$
		\STATE $R_i$ applies the zigzag search
		\ENDIF
		\STATE $t\gets t+\Delta t$
		\ENDWHILE
	\end{algorithmic}  
	\caption{soft-obstacles($\tau, r, \gamma, \Delta t$)}
	\label{algo:softobstacle}
\end{algorithm}

Regarding research issues, the soft obstacle strategy inherits completeness from the zigzag search and cellular decomposition \cite{IEEEexample:choset2001coverage}.
However, the soft obstacle strategy can suffer from high computational complexity. The soft obstacle strategy is based on cellular decomposition. Decomposition of exploration regions in the soft obstacle strategy application is computationally costly. In interaction, robots determine exploration regions to travel to.

Also, the coordination strategy is not recommended for situations where coordination involves a large number of robots, because the unknown environment might not be decomposed efficiently. If many robots have inaccessible regions, it might not be possible for 
each robot to acquire a non-overlapping region independently before the end of the search. Experiments on this aspect will be investigated in future.

\section{Results and Discussion}
\label{sec:experiments}

For experimental investigation, the accidental rendezvous strategy (ARS), the periodic rendezvous strategy (PRS) and our proposed soft obstacle strategy (SOS) are considered.

Two aspects will then be investigated: (1) Experimental investigation of the sustainability of the exploration performance of robots which start searching within the same vicinity. (2) Theoretical analysis of the most appropriate strategy for coordination of solitary robots. 
For the experimental investigation, we consider four ways to compare the three strategies (following the representation of \cite{IEEEexample:wellman2011using, IEEEexample:hourani2013serendipity}):
\begin{enumerate}[(i)]
	\item We run a one-shot experiment and we plot robot individual coverage at the end of the search.
	\item We show the robot coverage over time during the search.
	\item We run a one-shot experiment with each strategy and we show the robots' individual trajectories.
	\item We run $150$ experiments, $50$ each on three different environments, and plot means and standard deviations of coverage per strategy (Figure \ref{fig:error}).
\end{enumerate}

In terms of results, we expect ARS and SOS to outperform PRS in terms of coverage, because PRS suffers from interruptibility. On the other hand, PRS should outperform ARS and SOS in sustainability. We expect SOS to outperform ARS in both coverage and sustainability.
%
We use the standard deviations of robot coverages to measure sustainability. This should be small in the case of sustainable performance.
Performance will be evaluated based on coverage area. The maximum area that a robot can cover, $A(\tau)$, can be computed as in Equation \ref{eq:area}. Coverage will be reported as a percentage of $A(\tau)$. 
\subsection{Experimental setup}
\begin{figure}[h]
	\centering
	\begin{subfigure}[h]{0.1\textwidth}
		\includegraphics[width=\textwidth]{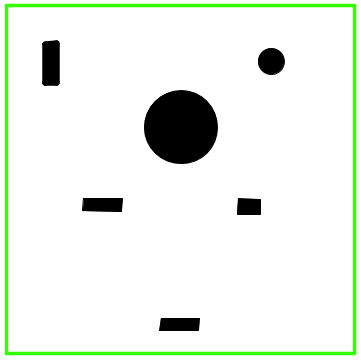}
		\caption{\scriptsize{}}
		\label{fig:world1}
	\end{subfigure}
	\begin{subfigure}[h]{0.1\textwidth}
		\includegraphics[width=\textwidth]{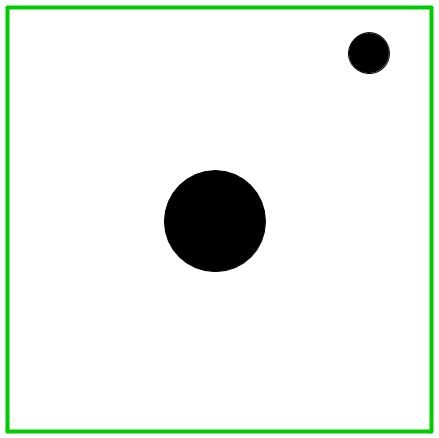}
		\caption{\scriptsize{}}
		\label{fig:world2}
	\end{subfigure}
	\begin{subfigure}[h]{0.1\textwidth}
		\includegraphics[width=\textwidth]{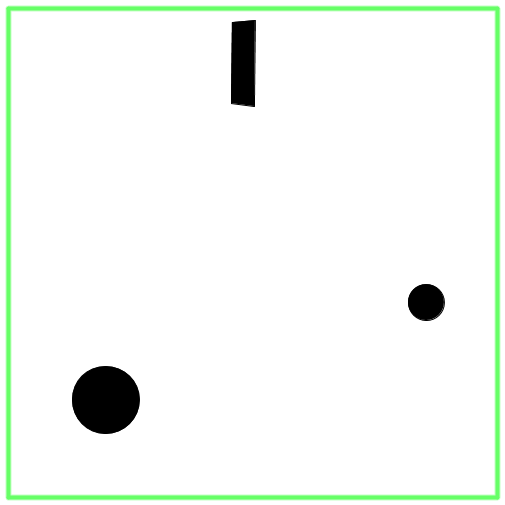}
		\caption{\scriptsize{}}
		\label{fig:world3}
	\end{subfigure}
	\caption{\small \textit{The simulation environments used.}}
	\label{fig:environment}
\end{figure}

The perception range was set to $r=20$ and the motion scale to $\gamma=1$. Three unstructured worlds were considered (Figure \ref{fig:environment}) and location-based sensing was simulated. The environment used to run simulations was $480\times 600$. The simulator was built in Python and Qt, a C++ cross-platform application framework for graphical user interfaces. We included our own implementation of ARS and PRS into the simulator.

In PRS, the rendezvous time $t_r$ was set as done in \cite{IEEEexample:hourani2013serendipity} as,
$$
t_r = 2\times a + b\,,
$$
where $a$ is the time required to travel between the rendezvous point and the current locations of the involved robots, and $b$ is a threshold which provides a time that robots can use to explore as they move to the rendezvous point. For the first rendezvous, $a$ is set to $50$ steps. For further rendezvous, it depends on the area of the shared explored regions between the robots involved. The rendezvous point is chosen by the leader in the shared explored region. 

The known upper bound amount of time for search is determined as follows,
$$\tau= k\times \bigg(\frac{ w h}{2\gamma rN}-\frac{\pi r}{2\gamma}\bigg)\,,$$
where $w$ and $h$ denote the width and the height of the search environment, and $k$ is defined in $ [0.5, 0.8]$. The factor $k$ is used to vary the width of the area to explore. Experiments were conducted (Table \ref{tab:experiments}) with $k=0.6$. 
\begin{table}[h]
	
	\centering
	\begin{tabular}{ |l|l|l|l|l|l|l|l|l| }
		\hline
		$N$&$2$&$3$&$4$&$7$&$8$&$10$\\
		\hline
		$\tau$&$2141$&$1421$&$1061$&$598$&$521$&$413$\\
		\hline
		$A$ &$86896$&$58096$&$43696$&$25176$&$22096$&$17776$\\
		\hline
	\end{tabular}
	\caption{\small \textit{Experimental setup: for each number of robots ($N$) and the time for search ($\tau$), the maximum possible coverage per robot $A(\tau)$ for this time will be used to evaluate robots' performance.}}
	\label{tab:experiments}
\end{table}
It should be noted that, this work does not define a process for determining environments to run experiments. Environments are randomly generated.
\subsection{Results}
\begin{figure}[h]
	\centering
	\begin{subfigure}[h]{0.24\textwidth}
		\includegraphics[width=\textwidth]{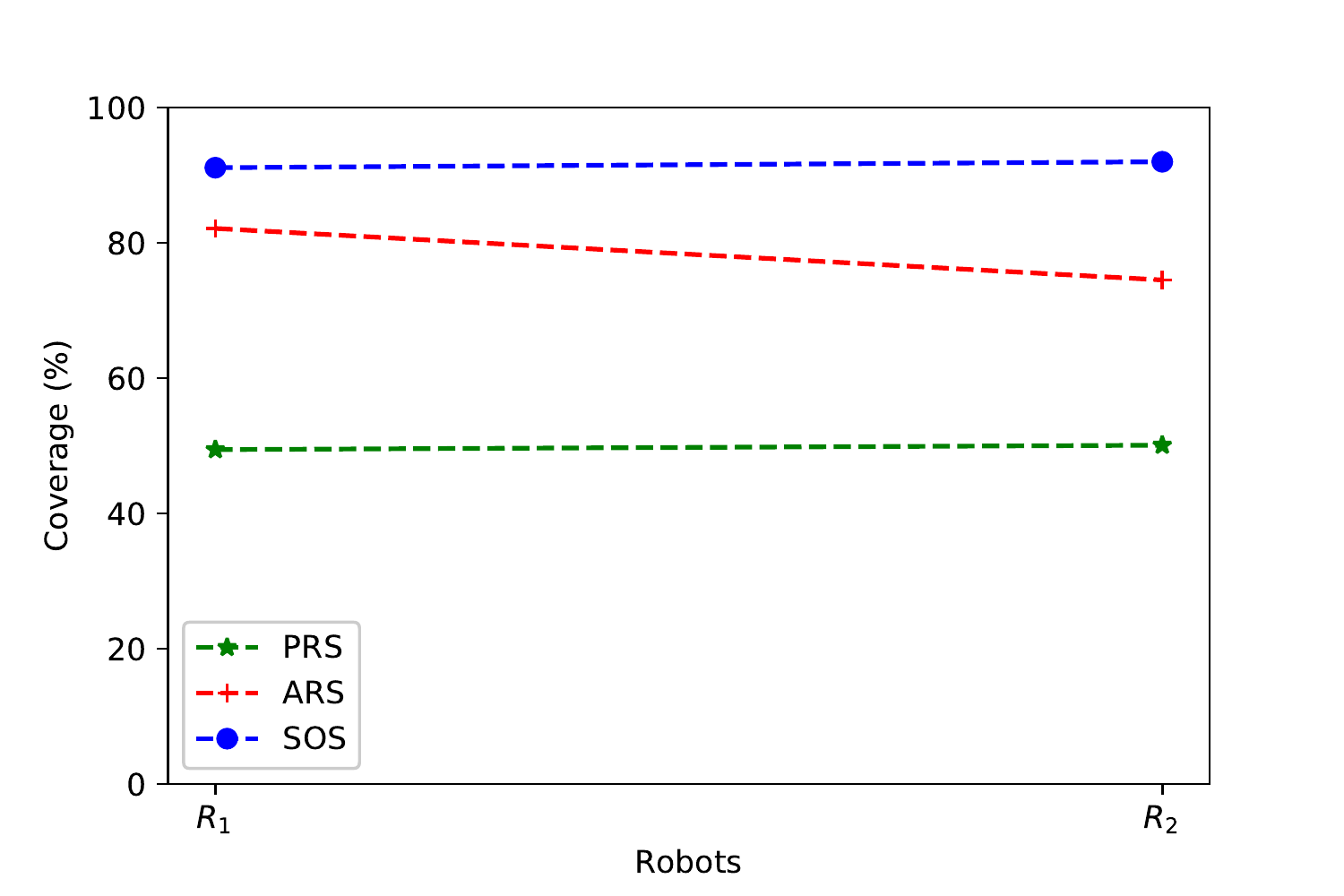}
		\caption{}
		\label{fig:indicov2}
	\end{subfigure}
	\begin{subfigure}[h]{0.24\textwidth}
		\includegraphics[width=\textwidth]{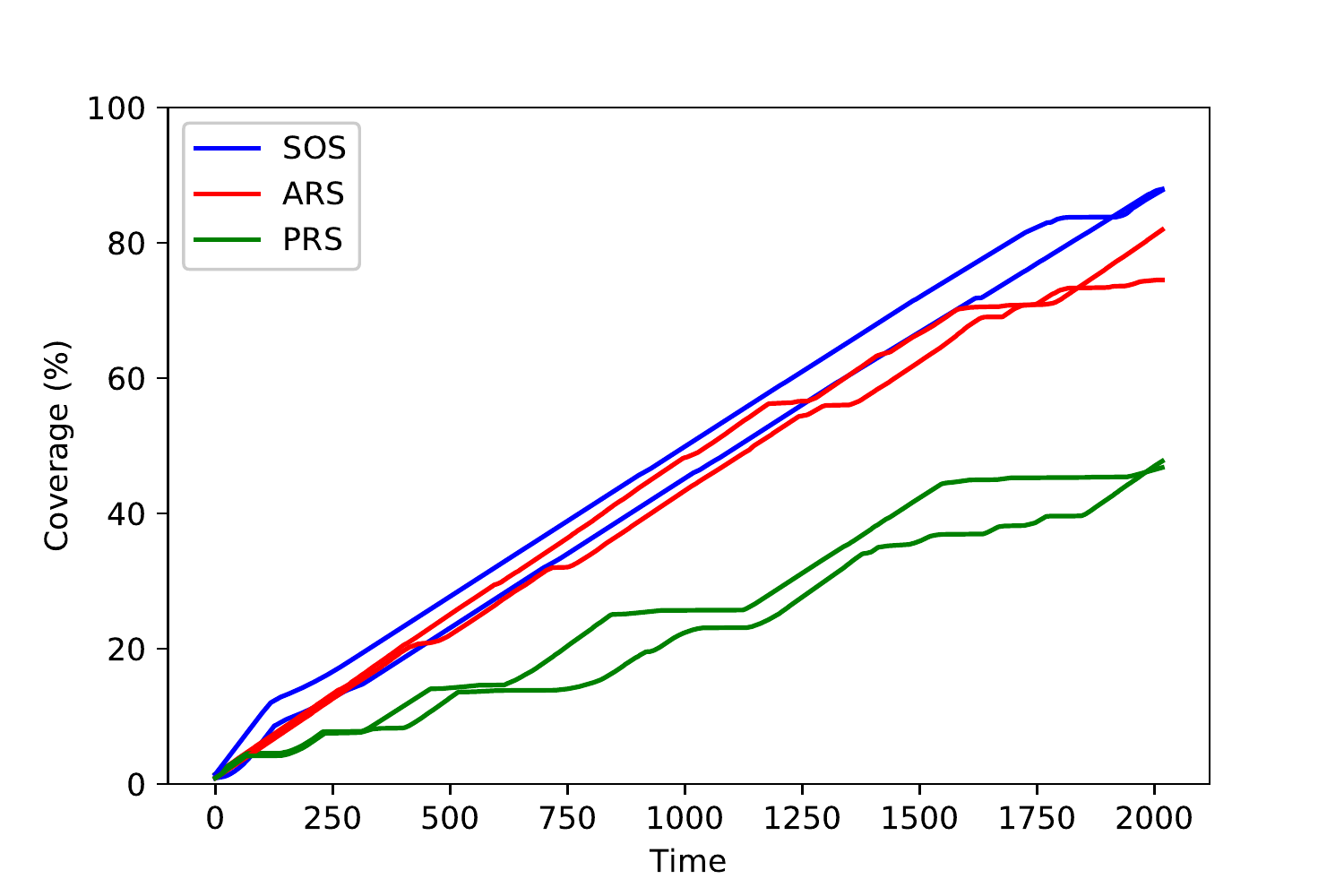}
		\caption{}
		\label{fig:progresscov2}
	\end{subfigure}
	\begin{subfigure}[h]{0.24\textwidth}
		\includegraphics[width=\textwidth]{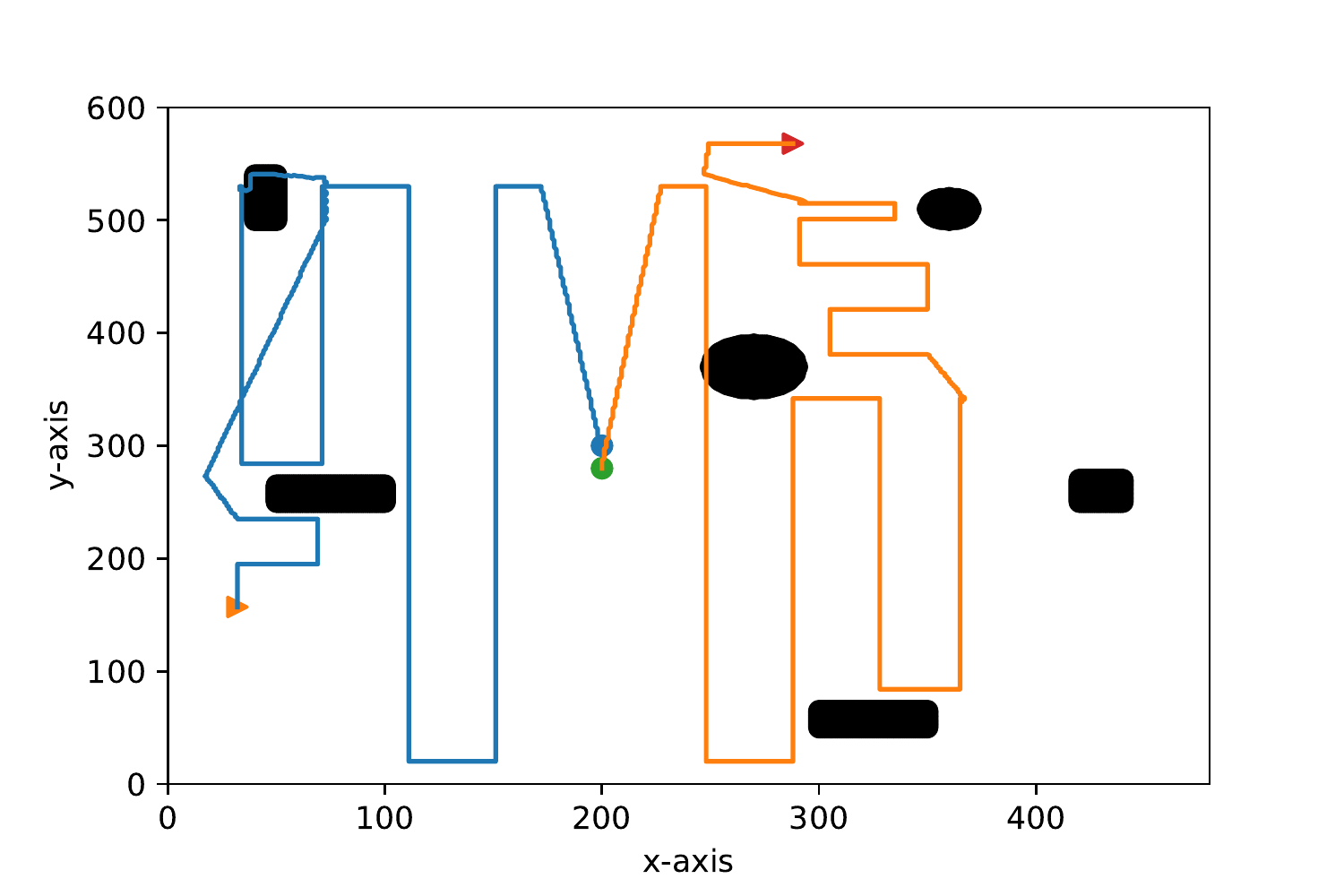}
		\caption{}
		\label{fig:sostraj2}
	\end{subfigure}
	\begin{subfigure}[h]{0.24\textwidth}
		\includegraphics[width=\textwidth]{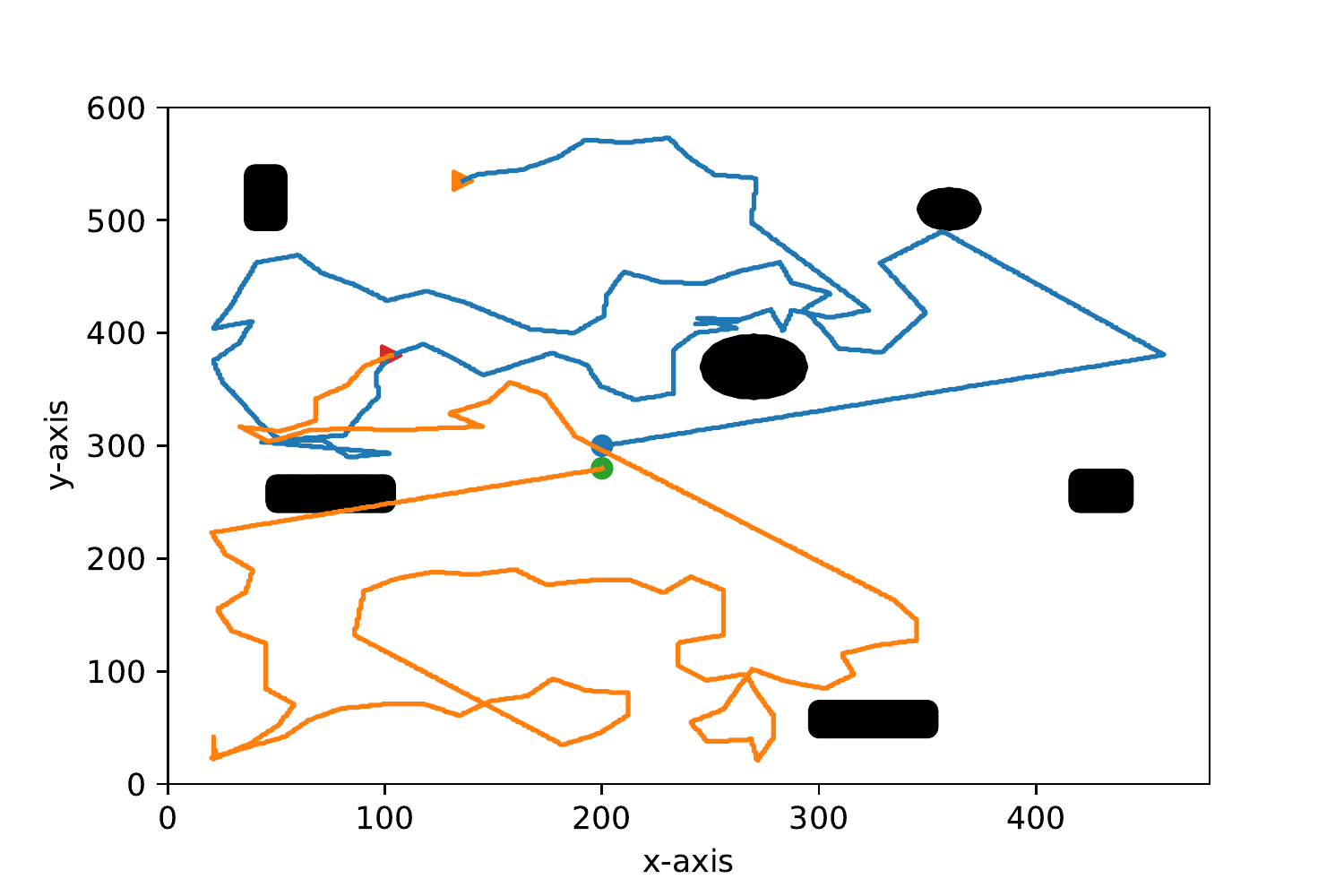}
		\caption{}
		\label{fig:arstraj2}
	\end{subfigure}
	\begin{subfigure}[h]{0.24\textwidth}
		\includegraphics[width=\textwidth]{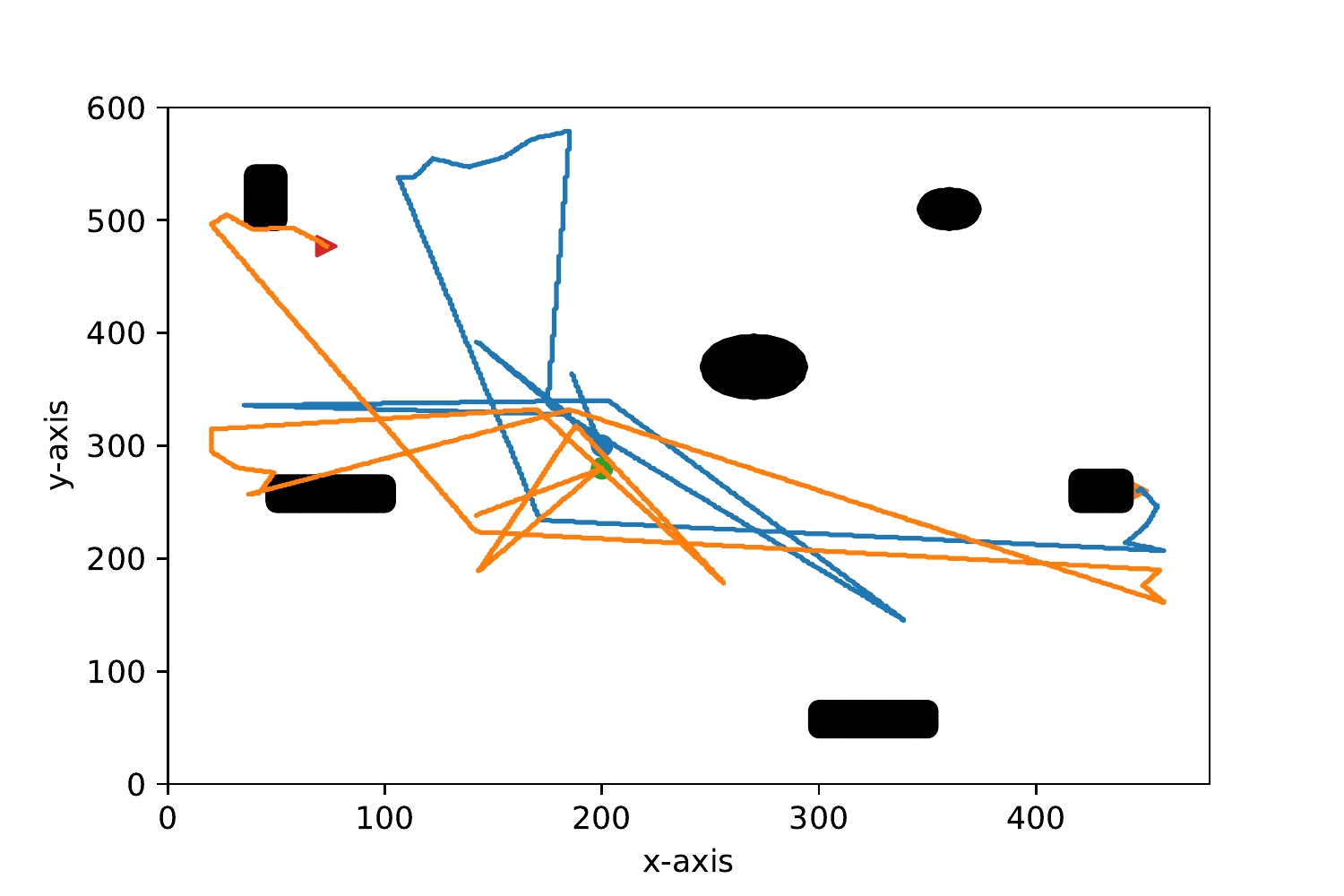}
		\caption{}
		\label{fig:prstraj2}
	\end{subfigure}
	\caption{\small \textit{Individual coverage performance for a team of two robots. \textbf{(\ref{fig:indicov2})}: final coverage achieved by individual robots. \textbf{(\ref{fig:progresscov2})}: coverage trends of coverage of individual robots over time. \textbf{(\ref{fig:sostraj2})}: trajectories of robots using SOS. Coloured dots are initial locations and coloured arrows are ending points of robot trajectories. \textbf{(\ref{fig:arstraj2})}: trajectories of robots using ARS. \textbf{(\ref{fig:prstraj2})}: trajectories of robots using PRS.}}
	\label{fig:experiment2}
\end{figure}

Figure \ref{fig:experiment2} shows the results for two robots in the unknown environment illustrated in Figure \ref{fig:environment}.
The three strategies perform well in terms of sustainability. The performance (in intervals in $\%$) of $A(\tau)$ of SOS, ARS and PRS are $87.857\pm 0.437$, $78.224\pm 3.715$ and $50.755\pm 0.326$ respectively. Robots had an even assignment. ARS and SOS have good performance in terms of coverage. But PRS has less coverage due to interruptibility which took roughly $49\%$ of the search time.

\begin{figure}[h]
	\centering
	\begin{subfigure}[h]{0.24\textwidth}
		\includegraphics[width=\textwidth]{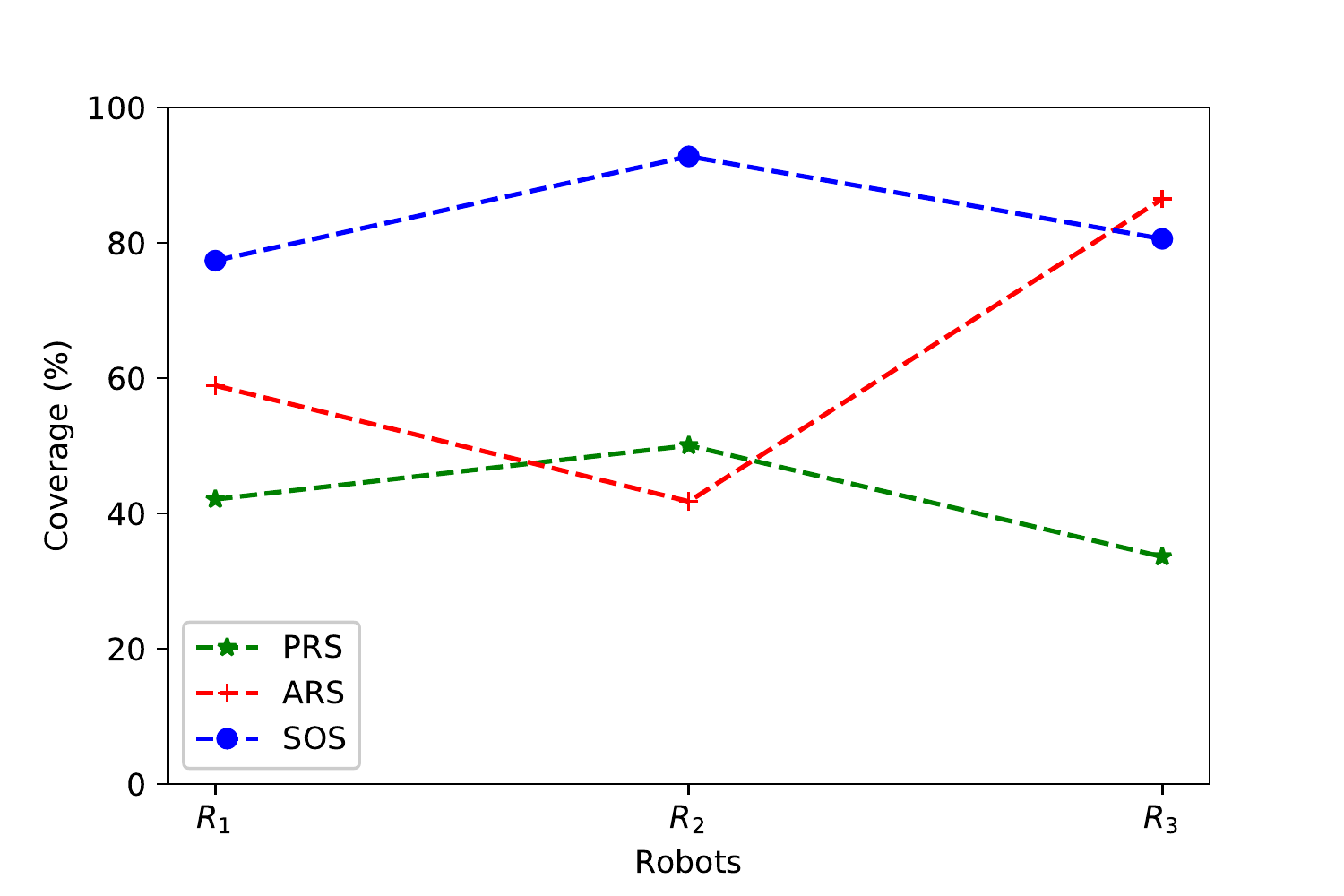}
		\caption{}
		\label{fig:indicov31}
	\end{subfigure}
	\begin{subfigure}[h]{0.24\textwidth}
		\includegraphics[width=\textwidth]{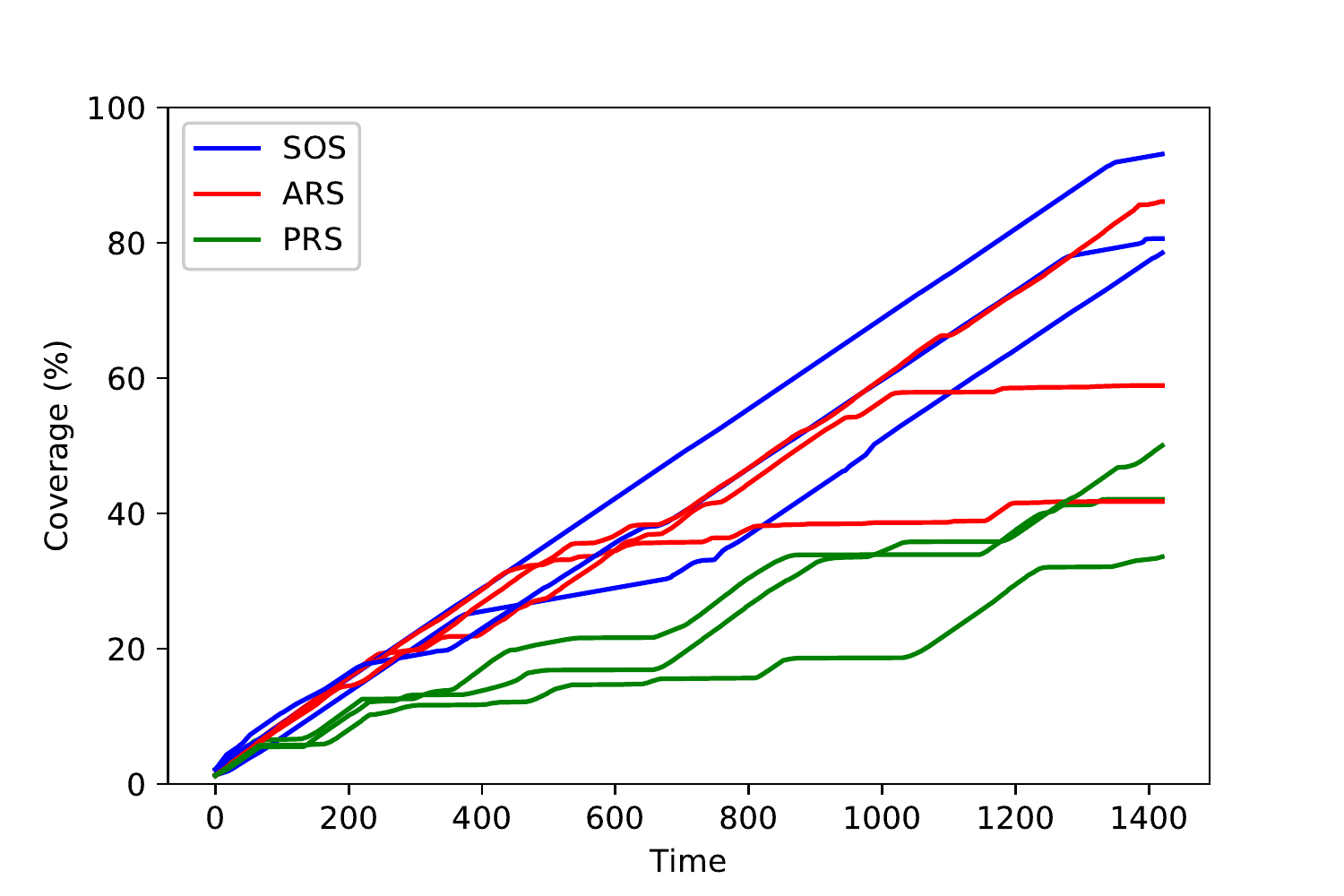}
		\caption{}
		\label{fig:progresscov31}
	\end{subfigure}
	\begin{subfigure}[h]{0.24\textwidth}
		\includegraphics[width=\textwidth]{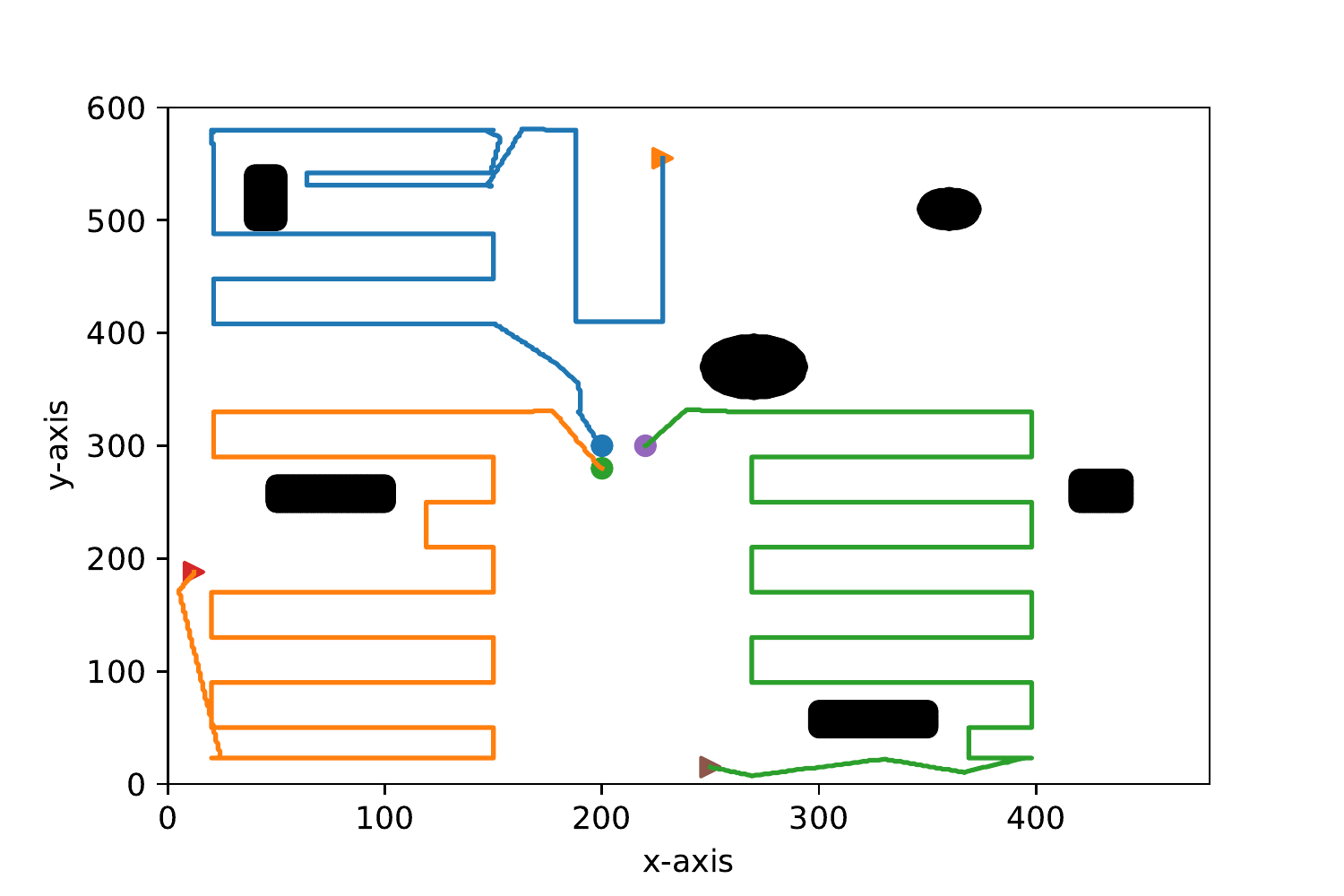}
		\caption{}
		\label{fig:sostraj31}
	\end{subfigure}
	\begin{subfigure}[h]{0.24\textwidth}
		\includegraphics[width=\textwidth]{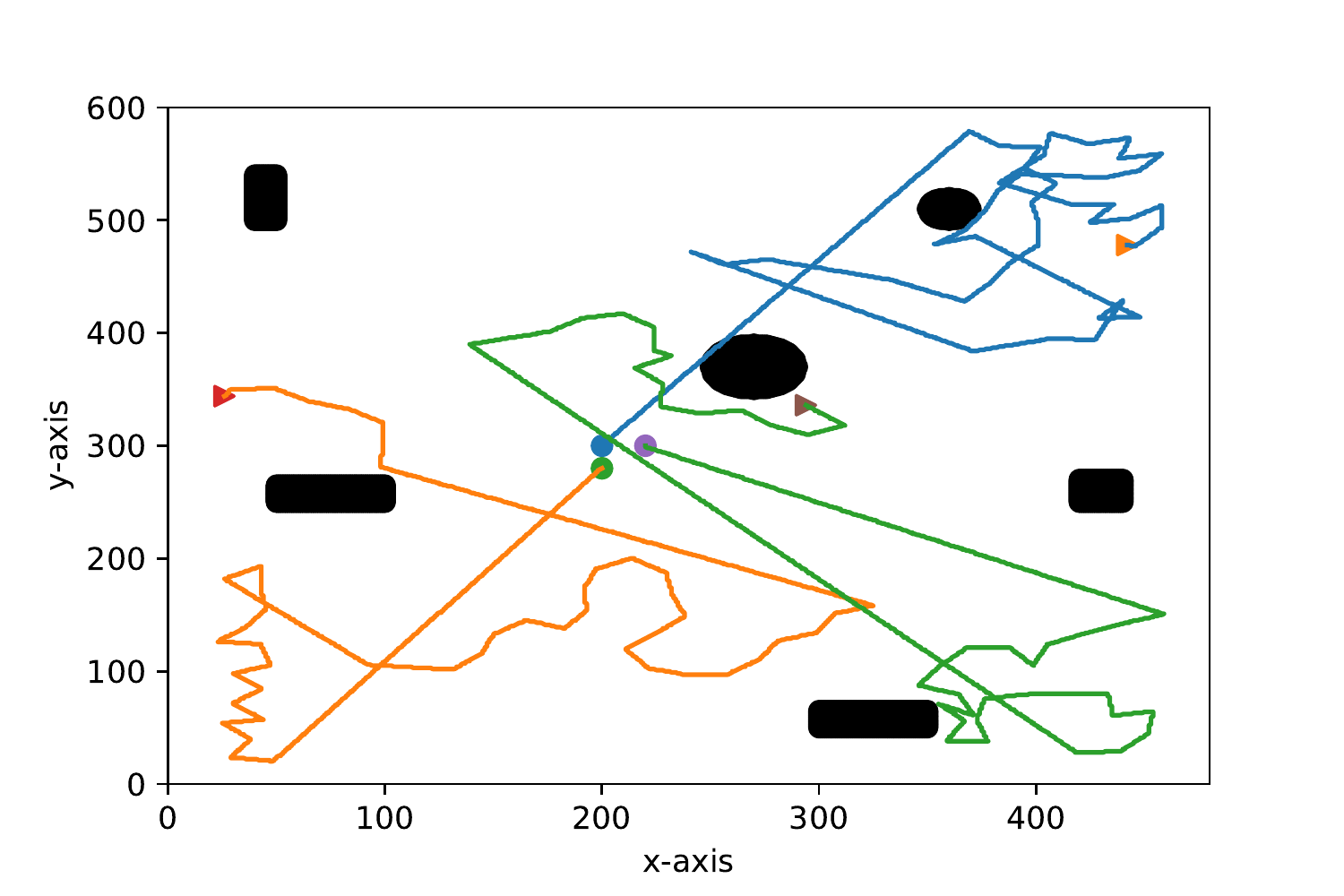}
		\caption{}
		\label{fig:arstraj31}
	\end{subfigure}
	\begin{subfigure}[h]{0.24\textwidth}
		\includegraphics[width=\textwidth]{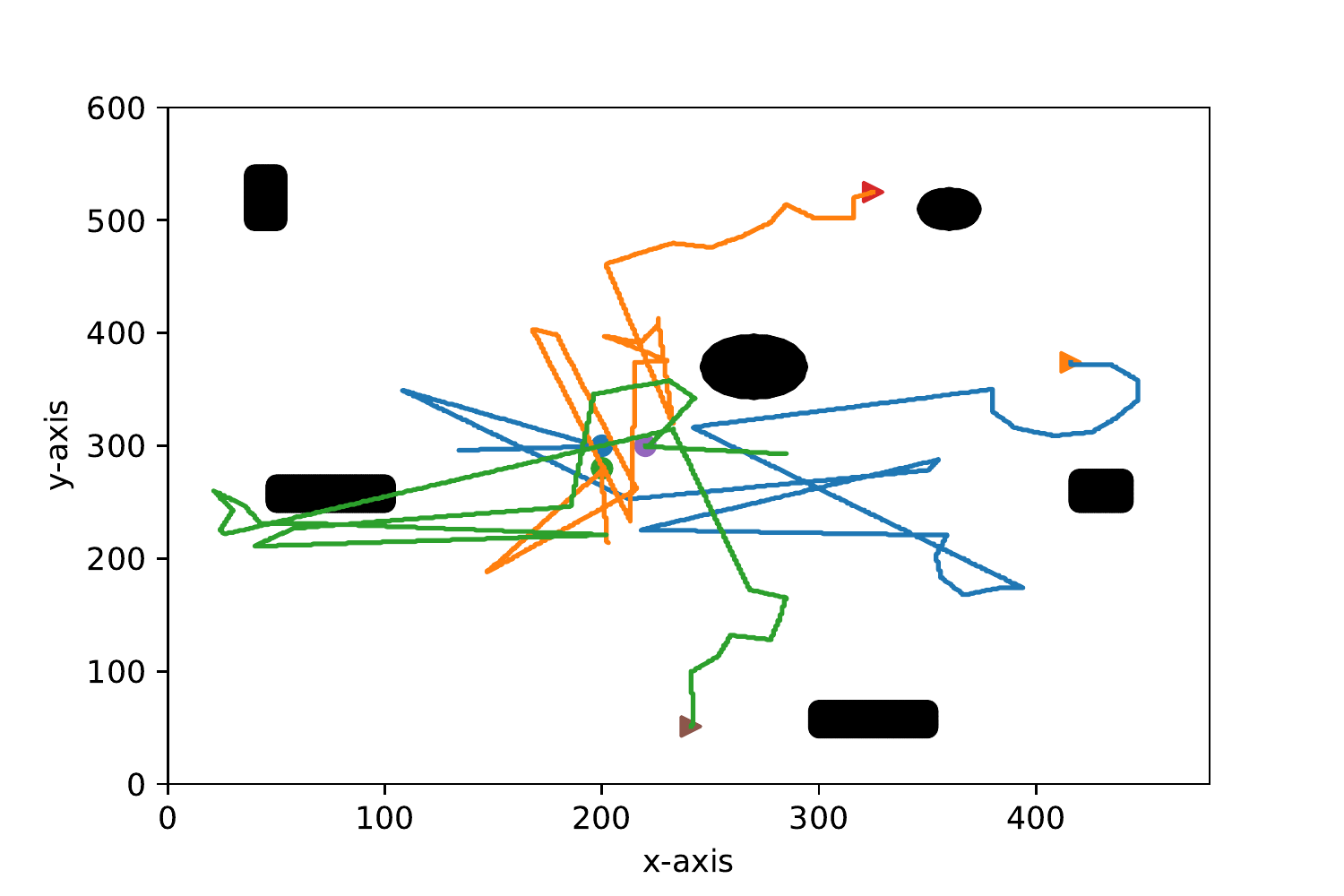}
		\caption{}
		\label{fig:prstraj31}
	\end{subfigure}
	\caption{\small \textit{Individual coverage performance for a team of three robots.}}
	\label{fig:experiment31}
\end{figure}

\begin{figure}[h]
	\centering
	\begin{subfigure}[h]{0.24\textwidth}
		\includegraphics[width=\textwidth]{figs/experiment32_coverage.pdf}
		\caption{}
		\label{fig:indicov32}
	\end{subfigure}
	\begin{subfigure}[h]{0.24\textwidth}
		\includegraphics[width=\textwidth]{figs/experiment32_progress.pdf}
		\caption{}
		\label{fig:progresscov32}
	\end{subfigure}
	\begin{subfigure}[h]{0.24\textwidth}
		\includegraphics[width=\textwidth]{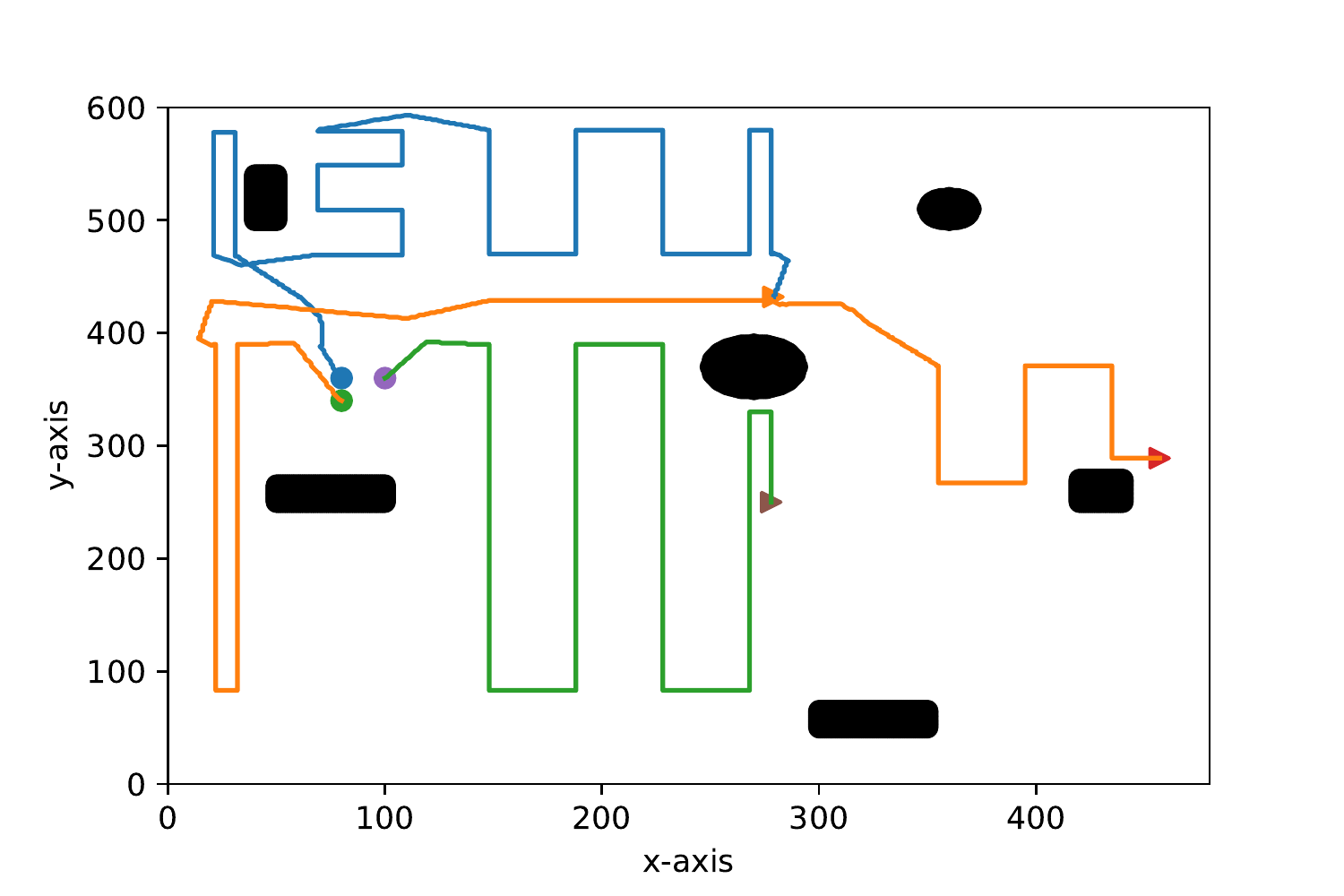}
		\caption{}
		\label{fig:sostraj32}
	\end{subfigure}
	\begin{subfigure}[h]{0.24\textwidth}
		\includegraphics[width=\textwidth]{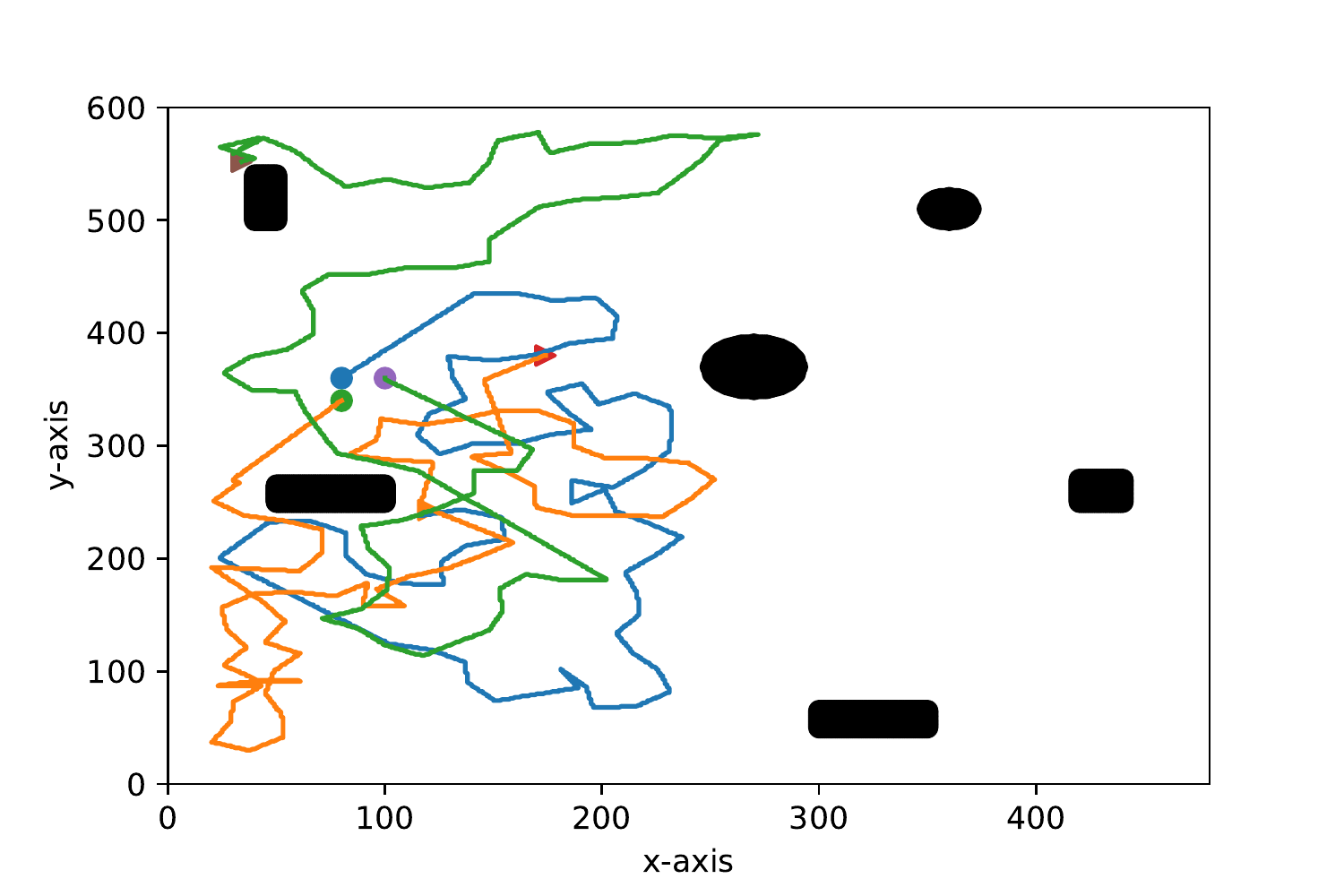}
		\caption{}
		\label{fig:arstraj32}
	\end{subfigure}
	\begin{subfigure}[h]{0.24\textwidth}
		\includegraphics[width=\textwidth]{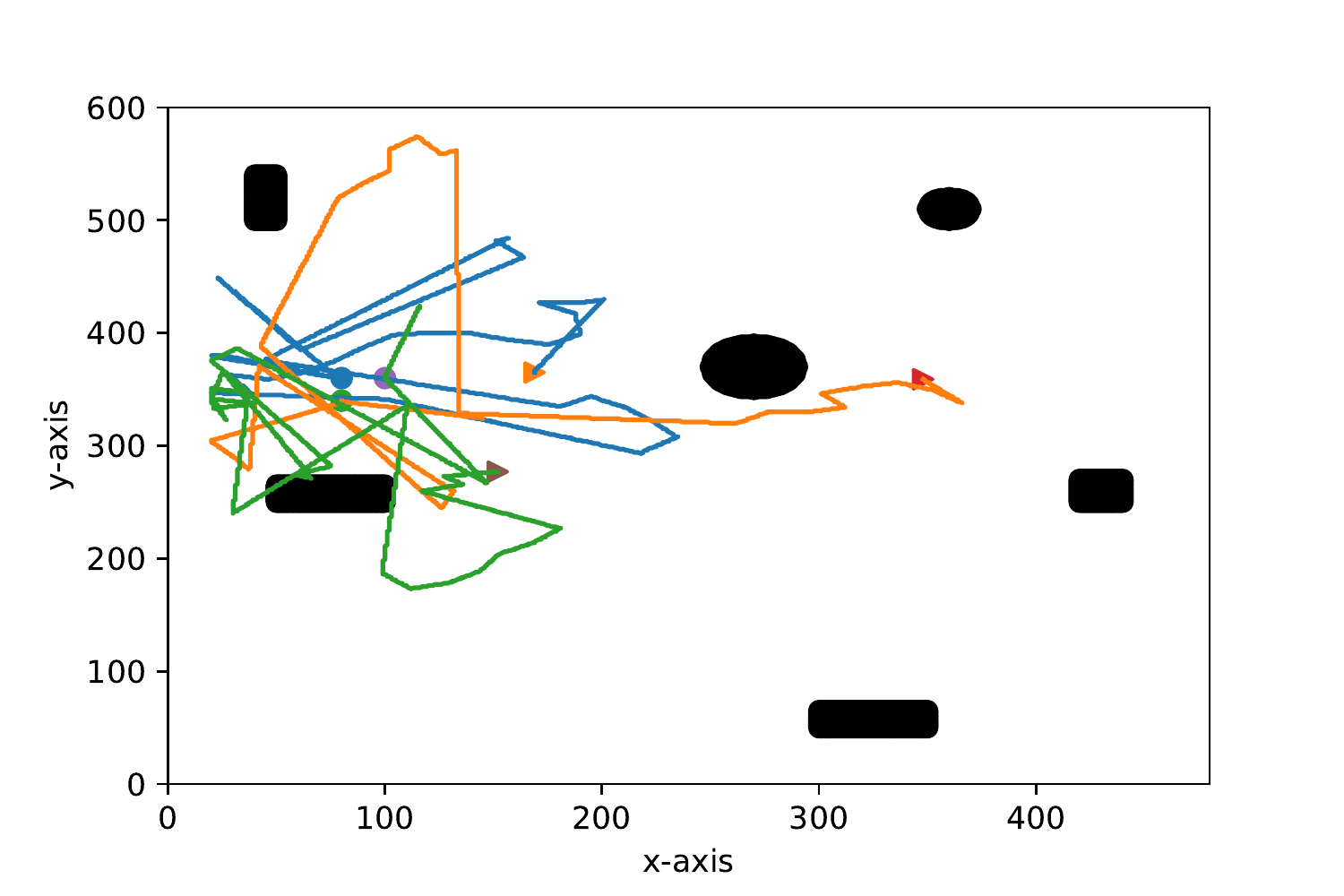}
		\caption{}
		\label{fig:prstraj32}
	\end{subfigure}
	\caption{\small \textit{Individual coverage performance for a team of three robots}.}
	\label{fig:experiment32}
\end{figure}

In Figure \ref{fig:experiment31}, robots also had an even assignment (the horizontal straight lines on PRS correspond to interruptibility), while Figure \ref{fig:experiment32} shows the results for an uneven assignment.
With ARS for the latter, two robots ($R_1$ and $R_3$) suffered from high interference. 

\begin{figure}[h]
	\centering
	\begin{subfigure}[h]{0.24\textwidth}
		\includegraphics[width=\textwidth]{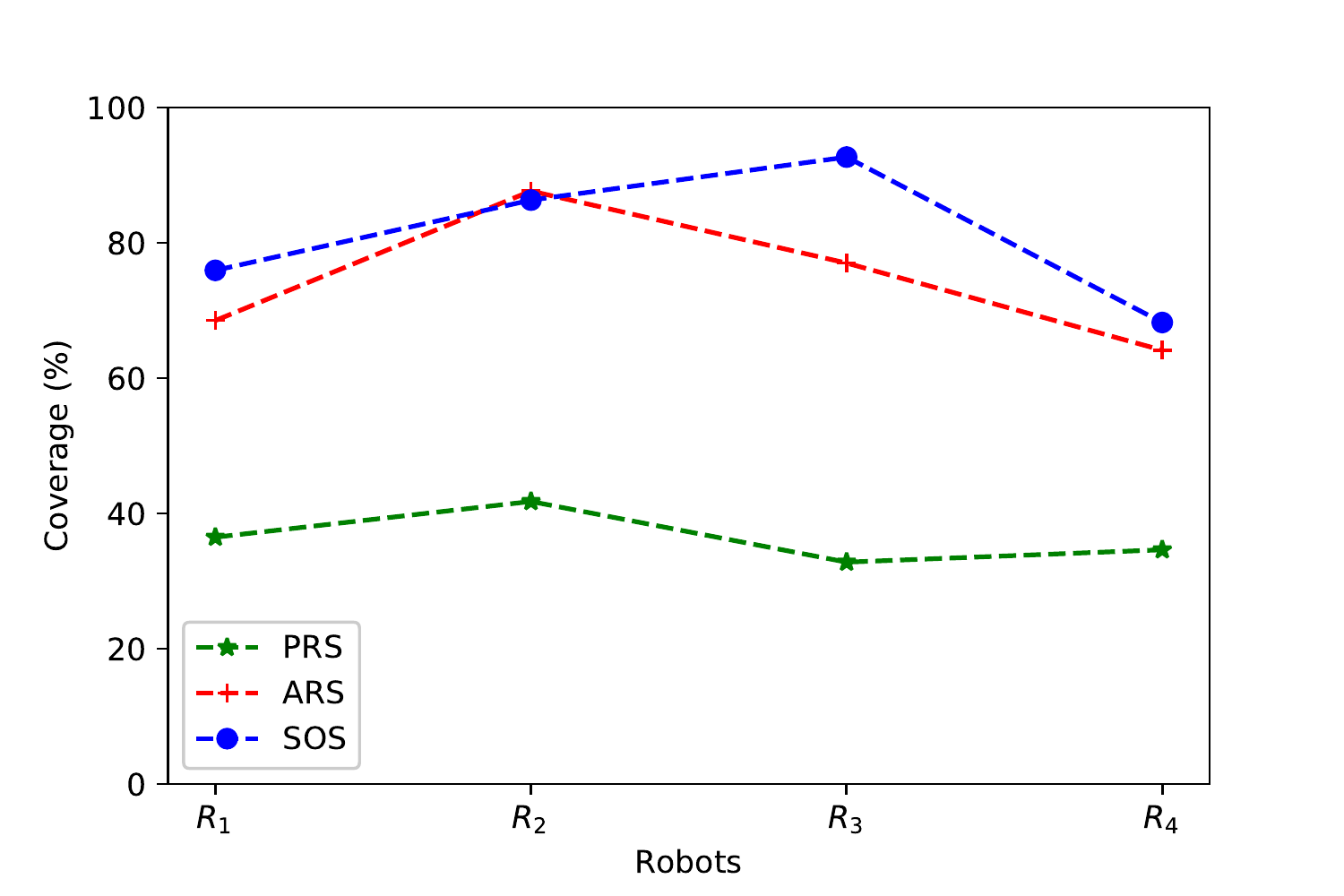}
		\caption{}
		\label{fig:indicov4}
	\end{subfigure}
	\begin{subfigure}[h]{0.24\textwidth}
		\includegraphics[width=\textwidth]{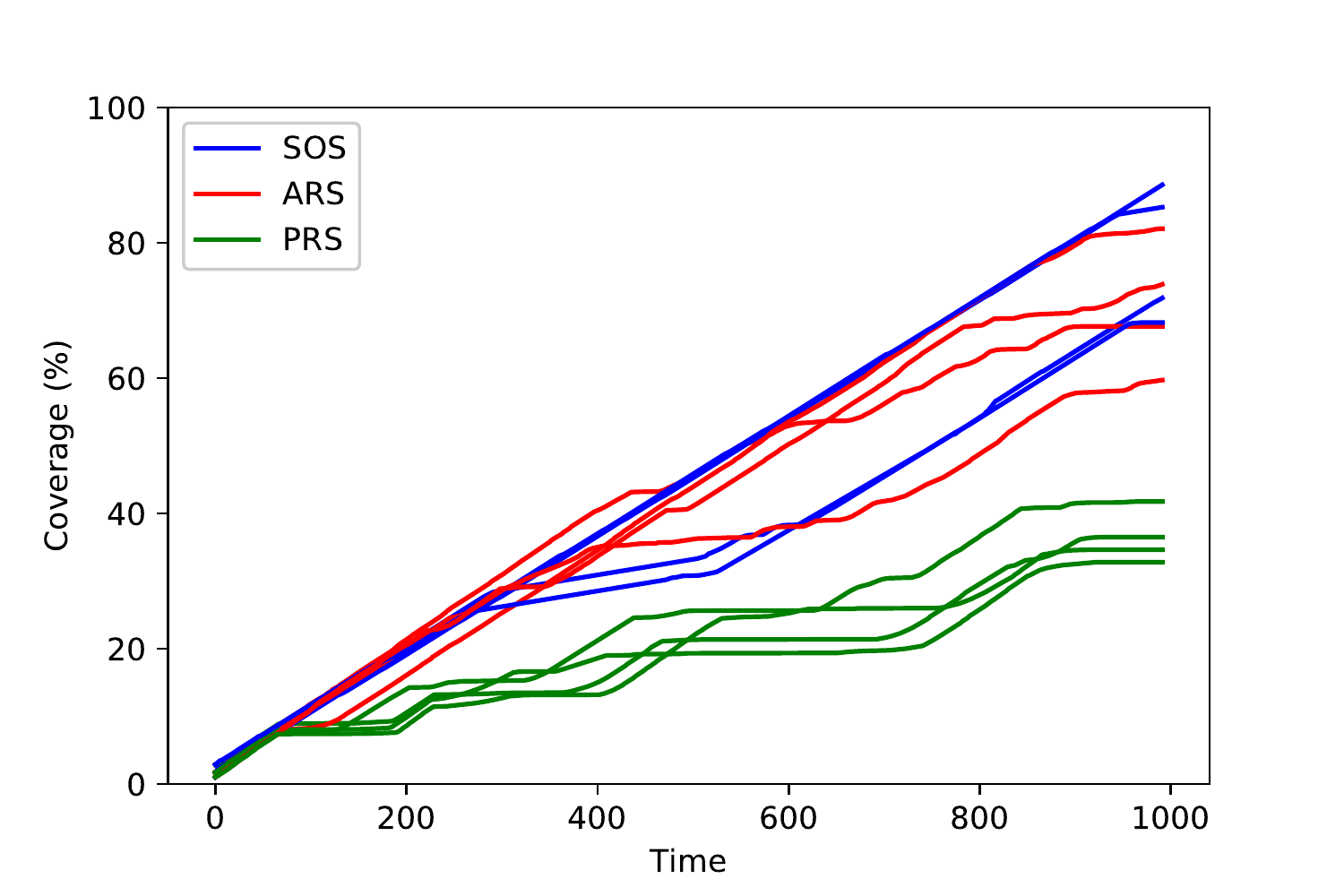}
		\caption{}
		\label{fig:progresscov4}
	\end{subfigure}
	\caption{\small \textit{Individual coverage performance for a team of four robots.}}
	\label{fig:experiment4}
\end{figure}
\begin{figure}[h]
	\centering
	\begin{subfigure}[h]{0.24\textwidth}
		\includegraphics[width=\textwidth]{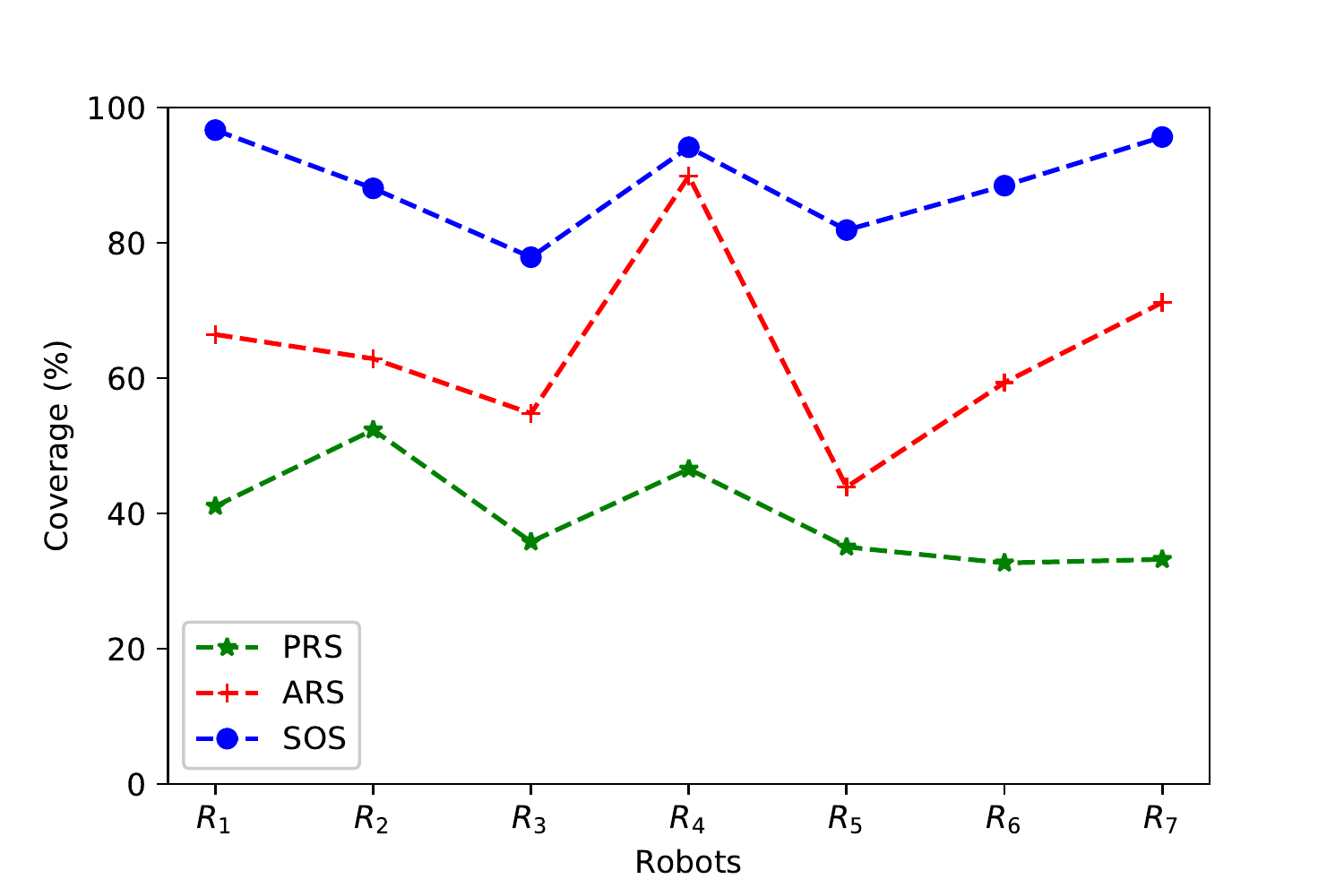}
		\caption{}
		\label{fig:indicov7}
	\end{subfigure}
	\begin{subfigure}[h]{0.24\textwidth}
		\includegraphics[width=\textwidth]{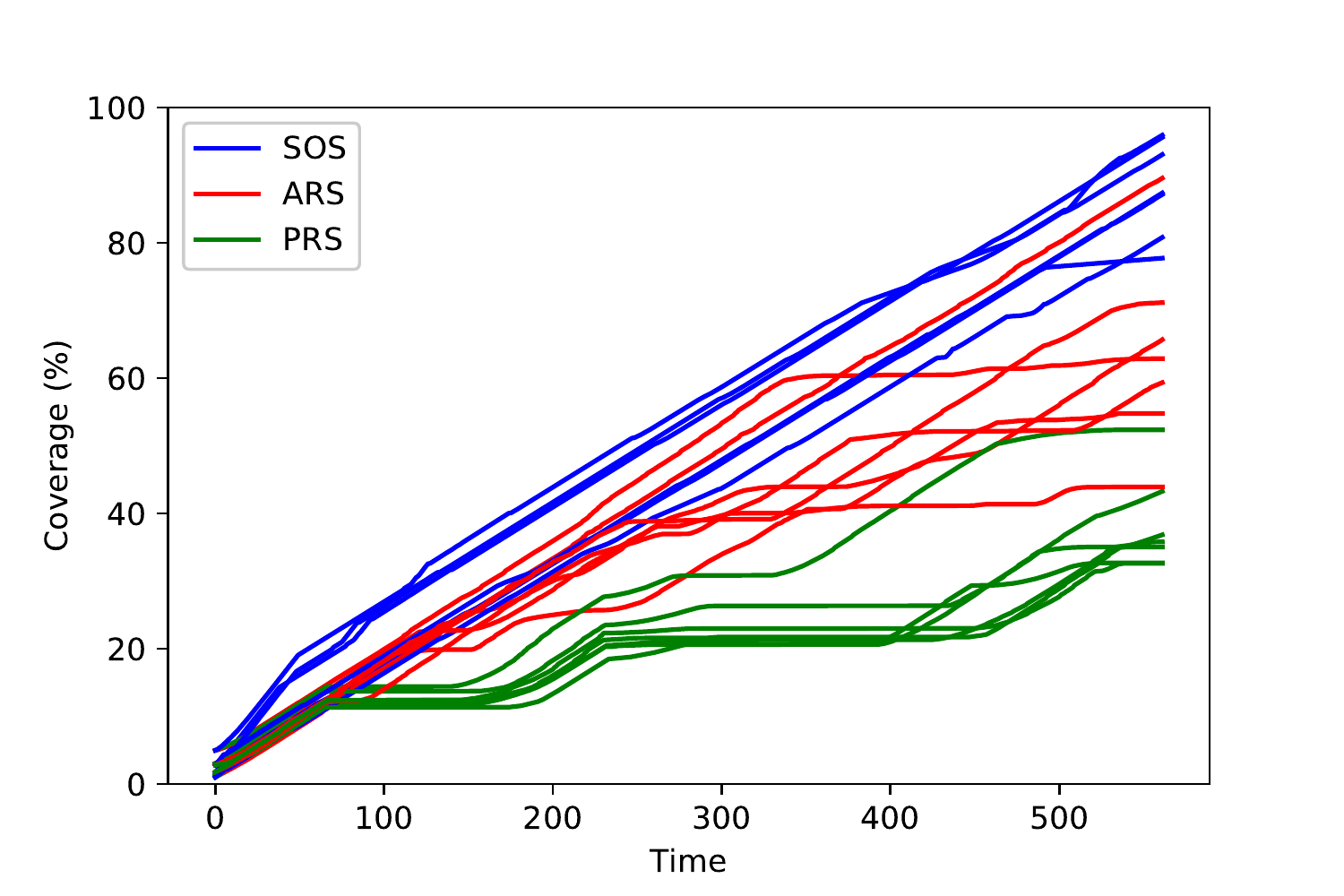}
		\caption{}
		\label{fig:progresscov7}
	\end{subfigure}
	\caption{\small \textit{Individual coverage performance for a team of seven robots}.}
	\label{fig:experiment7}
\end{figure}

\begin{figure}[h]
	\centering
	\begin{subfigure}[h]{0.24\textwidth}
		\includegraphics[width=\textwidth]{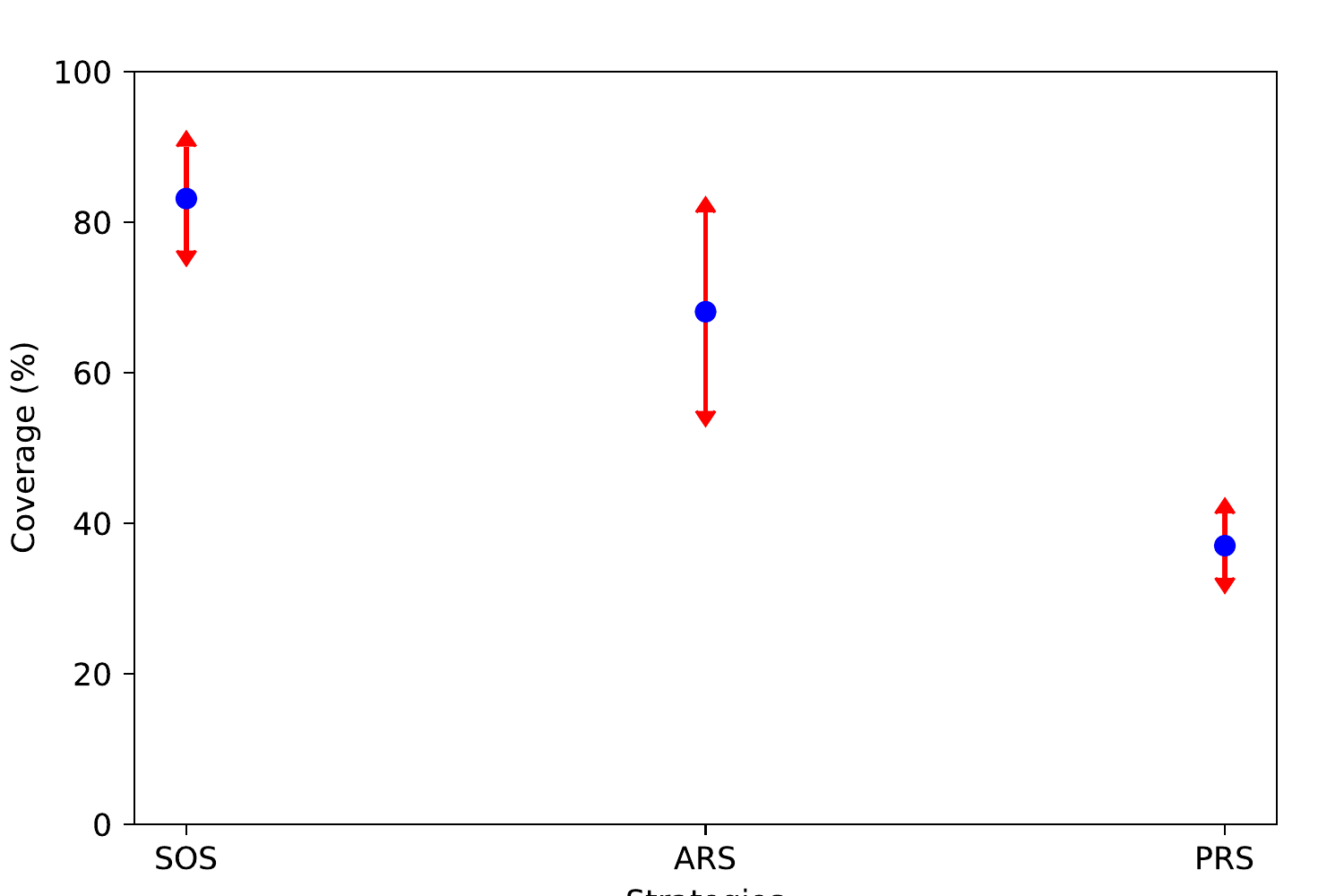}
		\caption{}
		\label{fig:mc8}
	\end{subfigure}
	\begin{subfigure}[h]{0.24\textwidth}
		\includegraphics[width=\textwidth]{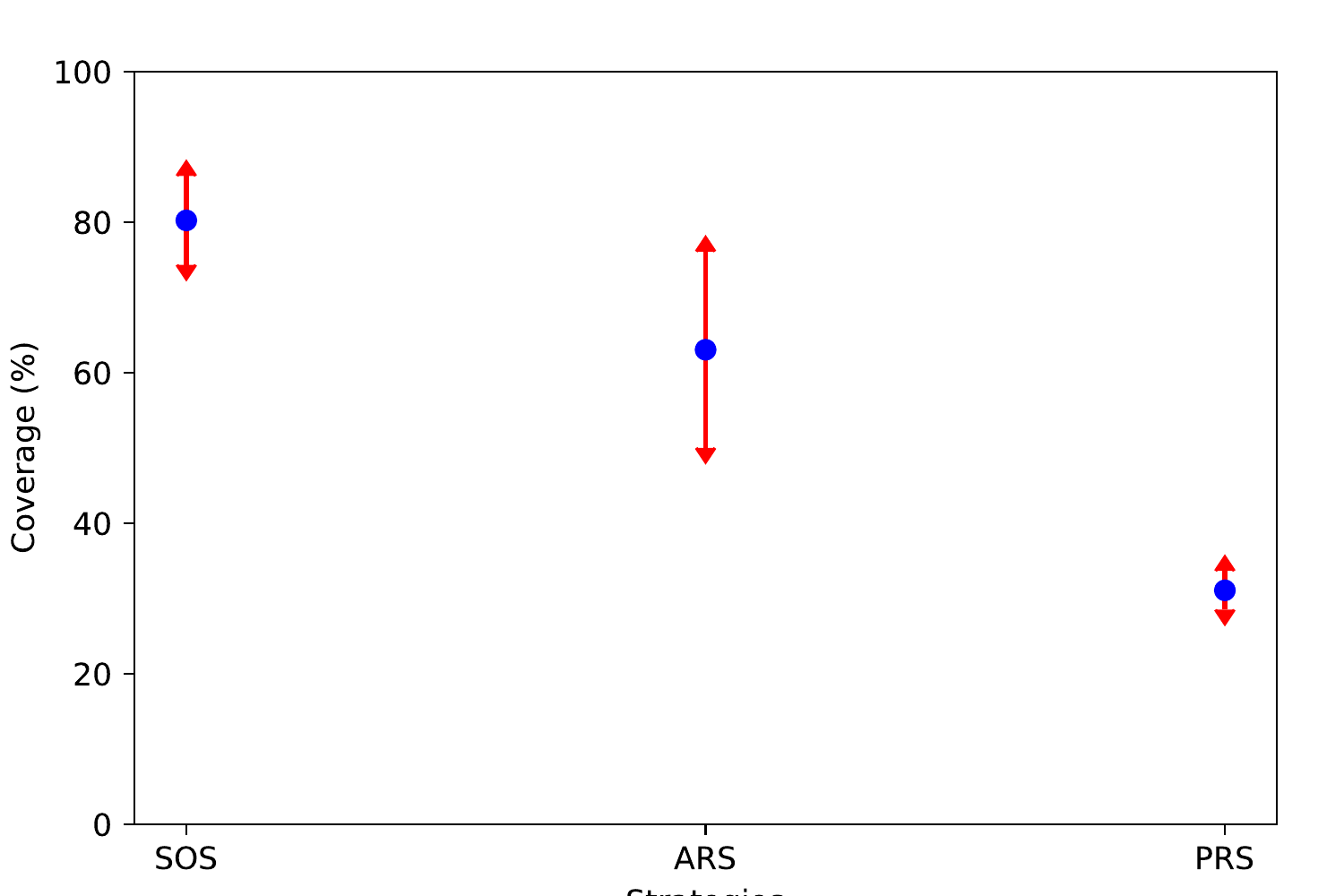}
		\caption{}
		\label{fig:mc10}
	\end{subfigure}
	\caption{\small \textit{Error bars for the coordination strategies' performance. Robots start from random locations in close proximity to each other. \textbf{(\ref{fig:mc8})}: error bars with a team of eight robots. \textbf{(\ref{fig:mc10})}: error bars with a team of ten robots. The normalised coverage (in $\%$) intervals obtained with SOS, ARS and PRS are $83.140\pm 6.913$, $68.108\pm 13.189$ and $37.016\pm 4.269$ for the eight robots respectively; and $80.239\pm 5.909$, $63.053\pm 13.046$ and $31.090\pm 2.574$ for the ten robots respectively.}}
	\label{fig:error}
\end{figure}

Figure \ref{fig:experiment4} and Figure \ref{fig:experiment7} show results for teams of four and seven robots respectively. In Figure \ref{fig:experiment4} robots had an uneven assignment. An uneven assignment does not affect the sustainability of PRS significantly because robots meet on a regular basis. Also ARS provides a good coverage performance ( $\approx 25\%$ of interference which includes overlap and backtracking). 
In Figure \ref{fig:experiment7},  $R_1$ has a low coverage as it was exploring an already explored region. As expected, PRS and SOS provide more sustainable performance than ARS.

Trajectories of robots (Figures \ref{fig:experiment2} -- \ref{fig:experiment7}) show that when SOS robots may explore non-overlapping regions. In the ARS case there is overlapping of regions.

Figure \ref{fig:error} shows summary coverage results from $150$ experiments. The results confirm that PRS provides the most sustainable performance (its standard deviation is very small) but with less coverage due to interruptibility. The robot performance obtained with SOS is more sustainable than that with ARS. SOS outperforms other strategies in coverage.
\subsection{Discussion}
\subsubsection{Sustainability of exploration performance}
the soft obstacle strategy works well when the search time is known in advance. Robots evaluate their exploration regions based on the amount of remaining search time. The case of unlimited time for search suits PRS the most.

ARS provides sustainable performance provided that robots have an even assignment in interaction \cite{IEEEexample:wellman2011using}. Otherwise, robots' explorations will overlap.
But sustainable performance is guaranteed in PRS \cite{IEEEexample:hourani2013serendipity}, since robots maintain collaboration through scheduled meetings. However, the approach is affected by interruptibility, which can hinder fast coverage. 

SOS can provide sustainable performance where the amount of search time is limited and known. While coordinating, the size of exploration regions assigned to robots by cellular decomposition is a function of the amount of time for search and the number of robots in interaction.

In the case of an uneven assignment with SOS, each of the robots concerned considers the regions of others as soft obstacles and avoids exploring these regions.

However, in the case of uneven assignment with a large number of interacting robots, SOS might not outperform the other two strategies. 
In addition, when considering a large number of robots, some robots may take long paths to reach their assigned regions. Consequently, these robots may spend more time traversing than searching.


\subsubsection{Applicability of the strategies for solitary robots}
PRS is used to manage a team of robots which can schedule meetings among themselves. In the case of solitary robots, a robot can encounter different robots at different times. For such robots to be incorporated into the team, a means would be required to schedule them into future rendezvous. Moreover, a robot encountering more than one team could have the problem of being a candidate for many other rendezvous. 
For instance, what happens if a robot which is going to rendezvous encounters another robot from another local group also going to rendezvous with its mates? If both decide to interact and share information, they might miss their respective meetings. If they do not interact, each misses the additional information from the other. Based on our knowledge, this issue is not yet fully addressed in the literature.

However, for situations such as map building where robots are bound by a team goal, ARS and SOS might not be recommended. Bound by a team goal, robots need to maintain collaboration. In ARS and SOS, collaboration of robots is always accidental.  Suppose a robot $R_j$ (that $R_i$ has previously met) meets another robot $R_k$ which $R_i$ is not aware of. How can $R_i$ learn from $R_k$? This situation can be nicely handled with PRS. Another potential difficulty of strategies which utilise accidental rendezvous, such as ARS and SOS, is that in some cases a large amount of time may be spent in these accidental interactions. This might particularly be so if there is a large number of robots relative to the coverage area.

%

\section{Conclusion}
\label{sec:conclusion}
A novel interruptibility-free strategy was designed to mitigate interference with sustainable performance in situations where solitary robots search an unknown environment when a known upper bound on the search time is insufficient to allow search of the entire search space. 

An improvement is observed when robots have an uneven assignment. An uneven assignment is when search areas assigned to robots are unbalanced. This can arise, for example, when robots ignore the fact that they meet close to a boundary of the search environment. 
However, the soft obstacle strategy is not recommended when robots are bound by a team goal (e.g. map construction) as robots need to maintain collaboration. 
In the periodic rendezvous strategy, exchange of individual information between robots is maintained. The soft obstacle strategy is
recommended for self-interested robots.

%


The soft obstacle strategy proposed focuses on a limited number of robots for coordination.  A further avenue of research could be to extend the work by considering how a large number of robots could apply the strategy for coordination.
\vskip 0.1in
\bibliographystyle{IEEEtran}
\bibliography{IEEEabrv,jordan_soft_obstacles}


\section*{Acknowledgements}
The authors would like to thank Prof. Jeff Sanders for fruitful exchanges and his sharp scientific opinions. 

S.W. Utete and J.F. Masakuna are members of the AIMS\footnote{African Institute for Mathematical Sciences (South Africa).} Research Centre which receives support from the NRF\footnote{National Research Foundation.}. This support is acknowledged with thanks.



%

\end{document}